\documentclass{article}

\PassOptionsToPackage{numbers,sort&compress,super,comma}{natbib}
\usepackage[preprint]{neurips_2026}
\usepackage{float}  
\usepackage{algorithm}
\usepackage{algpseudocode}
\usepackage{amsmath}
\usepackage{amssymb}
\usepackage{amsthm}

\usepackage{booktabs}
\usepackage{graphicx}
\graphicspath{{nc_images/}}
\usepackage{microtype}
\usepackage{xcolor}
\usepackage{helvet}  
\usepackage{makecell}
\usepackage{colortbl}
\usepackage{multirow}
\usepackage{threeparttable}
\definecolor{ncheader}{HTML}{E4E1C6}
\definecolor{ncrow}{HTML}{EEE7CA}
\usepackage{caption}
\DeclareCaptionLabelSeparator{ncbar}{~$\vert$~}
\usepackage{hyperref}
\usepackage[capitalise,noabbrev]{cleveref}
\usepackage{xr-hyper}
\usepackage{subcaption}
\usepackage{overpic}

\newcommand{\paneltitle}[2]{%
  \noindent{\sffamily\fontsize{10}{12}\selectfont #1.}\hspace{0.4em}%
  {\sffamily\fontsize{8}{10}\selectfont #2}\par\vspace{1pt}%
}
\newcommand{\sisec}[1]{Supplementary Section~\ref{#1}}
\newcommand{\sifig}[1]{Supplementary Fig.~\ref{#1}}
\newcommand{\sitab}[1]{Supplementary Table~\ref{#1}}

\title{Learning Responsibility-Attributed Adversarial Scenarios 
for Testing Autonomous Vehicles}

\author{%
\textbf{Yizhuo Xiao}\textsuperscript{1},\ \
\textbf{Haotian Yan}\textsuperscript{2},\ \
\textbf{Ying Wang}\textsuperscript{3},\ \
\textbf{Zhongpan Zhu}\textsuperscript{2,4},\ \
\textbf{Yuxin Zhang}\textsuperscript{5},\\
\textbf{Xintao Yan}\textsuperscript{6}, \ \ 
\textbf{Mustafa Suphi Erden}\textsuperscript{1},\ \
\textbf{Cheng Wang}\textsuperscript{1,$\ast$}\\
\textsuperscript{1}School of Engineering and Physical Sciences, Heriot-Watt University, Edinburgh, U.K.\\
\textsuperscript{2}State Key Laboratory of Autonomous Intelligent Unmanned Systems, \\ Tongji University, Shanghai, China\\
\textsuperscript{3}College of Computer Science and Technology, Jilin University, Changchun, China\\
\textsuperscript{4}University of Shanghai for Science and Technology, Shanghai, China\\
\textsuperscript{5}National Key Laboratory of Automotive Chassis Integration and Bionics, \\ Jilin University, Changchun, China\\
\textsuperscript{6}Department of Civil Engineering, The University of Hongkong, Hongkong, China\\
\textsuperscript{$\ast$}Corresponding author: \texttt{cheng.wang@hw.ac.uk}
}

\makeatletter
\renewcommand{\@biblabel}[1]{#1.}
\makeatother

\begin{document}
\maketitle

\begin{abstract}
\noindent
Establishing trustworthy safety assurance for autonomous driving systems (ADSs) requires evidence that failures arise from avoidable system deficiencies rather than unavoidable traffic conflicts.
Current adversarial simulation methods can efficiently expose collisions, but generally lack mechanisms to distinguish these fundamentally different failure modes.
Here we present CARS (Context-Aware, Responsibility-attributed Scenario generation), a framework that integrates responsibility attribution directly into adversarial scenario generation.
CARS combines context-aware adversary selection with a generative adversarial policy optimized in closed-loop simulation to construct collision scenarios that are both physically feasible and diagnostically attributable.
Across benchmark datasets spanning heterogeneous national traffic environments, CARS consistently discovers feasible collision scenarios with high attribution rates under multiple regulation-prescribed careful and competent driver models.
By coupling adversarial generation with normative responsibility assessment, CARS moves simulation testing beyond collision discovery toward the construction of interpretable, regulation-aligned safety evidence for scalable ADS validation.
\end{abstract}

\section{Introduction}
\label{sec:intro}

Autonomous driving systems (ADSs) are moving toward broader public-road testing and deployment.
Regulators, developers, and the public therefore need credible evidence that these systems can handle rare, high-risk traffic interactions~\cite{qian2026test}.
Road testing alone cannot provide that evidence at practical scale.
Safety-critical events are rare in naturalistic traffic, and statistical confidence from public-road exposure would require extremely large driving distances~\cite{kalra2016driving, feng2023dense, liu2024curse}.
Closed-loop simulation and scenario-based testing have therefore become central to ADS safety validation~\cite{feng2021intelligent, yan2023learning, ding2023survey, qian2026test}.
These methods are also increasingly connected to regulatory practice and to evaluation-efficient use of test evidence~\cite{unece2021r157, tang2024scenario, song2025synthetic, wu2026make}.

Recent work has made generated safety-critical scenarios more frequent, severe, diverse, controllable, or reactive to the ADS~\cite{rempe2022generating, chang2024safe, xu2025diffscene, zhang2023cat, xie2024real, zhu2025critical, xiao2025cld, wang2025had}.
These advances are important.
However, most adversarial generators still optimize or report collision occurrence as the primary outcome.
A collision count alone cannot distinguish ADS failures from unrealistic or infeasible adversarial motion, or from encounters that no human driver could reasonably avoid. For instance, a crash caused by an unavoidable sudden intrusion cannot meaningfully be attributed to deficient ADS decision-making.
We argue that inducing the generation of scenarios with elevated collision rates is a necessary condition for adversarial scenario generation, but the core criterion for evaluating test validity is whether a collision can be attributed to the system under test. Existing adversarial scenario generation methods treat collision itself as the optimization objective, conflating collisions that the system should have avoided but did not with collisions that no reasonable driving behavior could have prevented, which fundamentally undermining the credibility of test conclusions.

\begin{figure}[!htbp]
    \centering
    \includegraphics[width=\textwidth]{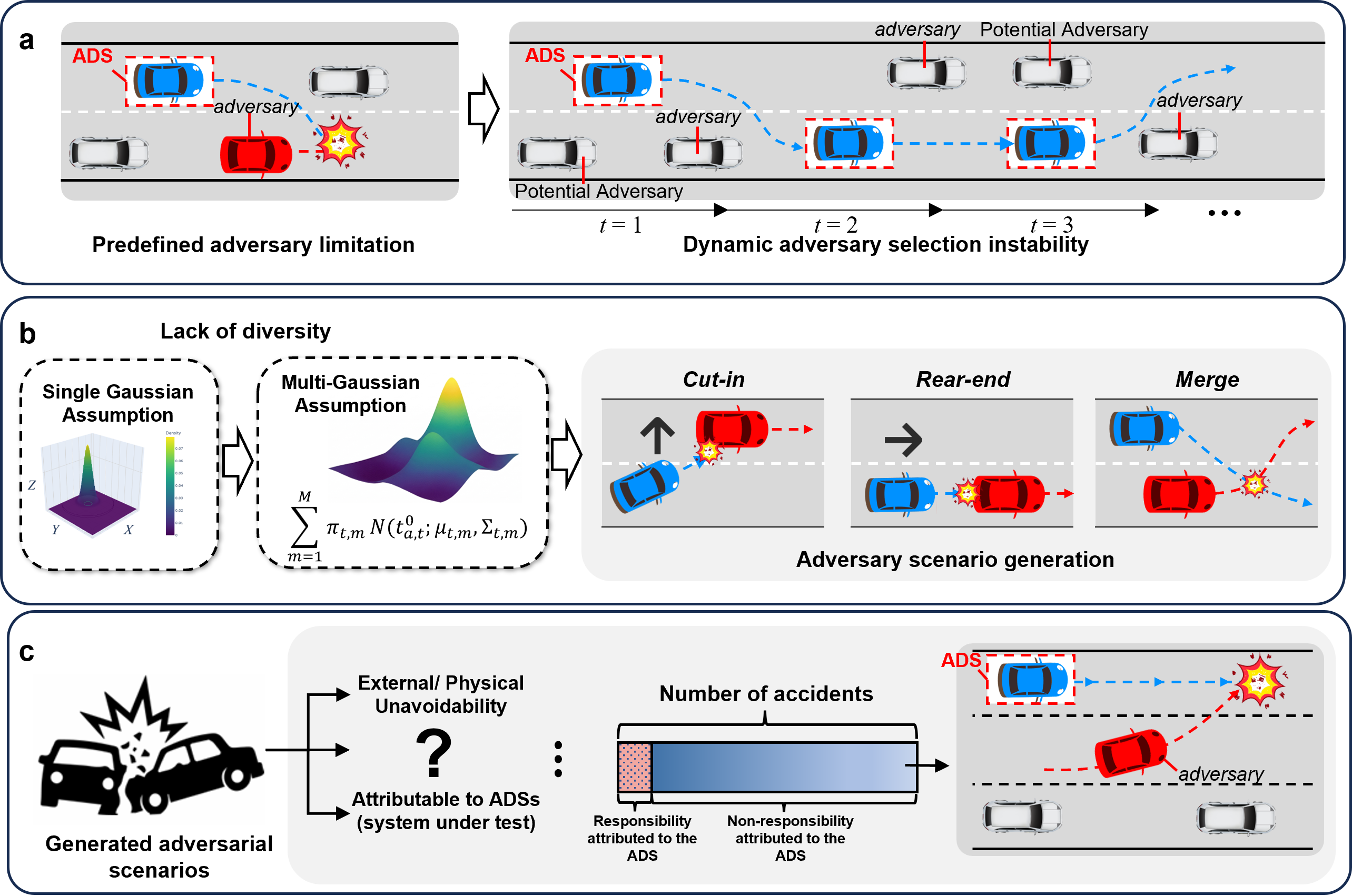}\\[4pt]
      \caption{Conceptual challenges in responsibility-attributed adversarial scenario generation.
      \textbf{a}, \emph{Adversary selection instability.} Static assignment or one-shot geometric heuristics fix the attacker (\textcolor{red}{red}) at simulation start; as traffic evolves the original adversary drifts out of the conflict region while other agents (\textcolor{gray}{gray}) that develop a credible threat are never reconsidered, leaving the ADS (\textcolor{blue}{blue}) unopposed.
      \textbf{b}, \emph{Trajectory-capacity limitation.} A policy that predicts a single-Gaussian distribution can still generate severe collisions, but the resulting severity may not survive attribution and feasibility checks.
      \textbf{c}, \emph{Responsibility ambiguity.} Without a responsibility-attribution standard, scenarios in which the ADS could have braked in time and scenarios that any reasonable response would also have suffered are reported under the same failure-rate metric, conflating ADS shortcomings with the encounter's inherent unavoidability.}
    \label{fig:challenges}
\end{figure}

Resolving responsibility attribution requires a behavioral reference standard that is independent of the system under test, used to delineate the encounters an ADS should reasonably be expected to avoid. Without such a reference, the same collision count mixes failures of the ADS with collisions that no reasonable driver could have prevented, and the resulting test evidence cannot be interpreted as evidence of ADS deficiency. The international regulatory framework UN ECE\,R157~\cite{unece2021r157} already provides the conceptual basis for such a standard through the careful and competent driver model (CCDM), a normative reference that defines the response a careful and competent driver (CCD) would produce in the same encounter. A collision is attributable to the ADS under test if the CCDM would have avoided it under the same scenario; otherwise the encounter lies outside the scope of meaningful ADS attribution. The CCD concept has been instantiated as several computational models with different structural assumptions, including the Fuzzy Safety Model (FSM)~\cite{mattas2020fuzzy, mattas2022fuzzy} prescribed by UN ECE\,R157, the Japanese CCD model (CC-JP)~\cite{jama2020framework}. A related formalism, Responsibility-Sensitive Safety (RSS)~\cite{ShalevShwartz2017OnAF} instead defines safety envelopes. These models share the same attribution rule and differ mainly in how they bound the reference driver's braking response and reaction delay.
  
However, existing adversarial scenario generators are not organized around this attribution rule. Their optimization remains at the level of inducing collisions. This leaves adversarial scenario generation with a structural gap, which spans adversary selection, trajectory generation, and collision evaluation, as summarized in \Cref{fig:challenges}. 
A responsibility-attributed scenario must begin with an adversary that can create an avoidable conflict from a plausible traffic position.
Its generated motion must remain feasible while exerting collision pressure.
Most importantly, the generation target should be an ADS-attributable collision: one that occurs for the ADS under test but would be avoided by a CCD reference.
These requirements link adversary selection and trajectory generation to the CCD attribution reference.

\begin{figure}[!htbp]
    \centering
    \includegraphics[width=\textwidth]{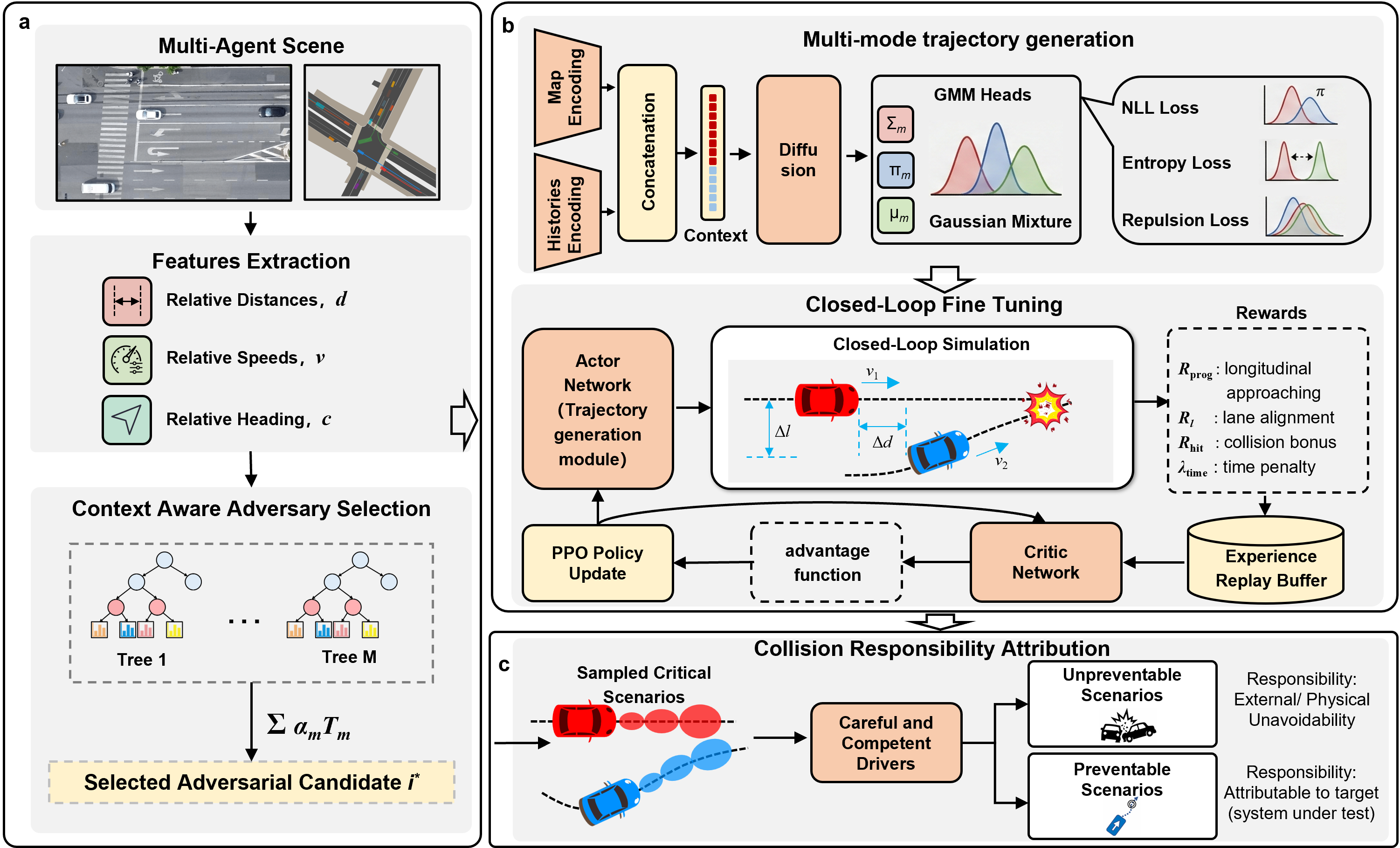}\\[4pt]
    \caption{CARS responsibility-attributed adversarial scenario generation framework.
\textbf{a}, \emph{Context-aware adversary selection.}
At each simulation step, CARS scores surrounding agents by their kinematic relationship to the ADS and stabilizes the selected agent over time, allowing the threat assignment to track the evolving conflict.
\textbf{b}, \emph{Adversarial trajectory generation.}
The selected agent is initialized from a Gaussian-mixture diffusion prior.
Closed-loop reinforcement learning fine-tunes the agent to approach the ADS and create safety-critical interactions.
\textbf{c}, \emph{Attribution-aware objective.}
CARS targets ADS-attributable collisions under a CCD reference.
A collision is attributable when it occurs for the ADS under test but would have been avoided by the CCD in the same encounter.
}
\label{fig:framework}
\end{figure}

To address this gap, we present CARS (\textbf{C}ontext-\textbf{A}ware, \textbf{R}esponsibility-attributed \textbf{S}cenario generation), a framework for generating ADS-attributable adversarial scenarios under a CCDM reference (\Cref{fig:framework}).
CARS is built around three design properties.
First, \emph{context-aware selection} chooses the adversarial counterpart from agents that can create a genuine avoidable conflict from their current traffic position.
Second, \emph{multi-component generation} uses a Gaussian-mixture diffusion policy to represent multiple plausible action-sequence hypotheses under the same traffic context~\cite{chen2025gaussian, kim2025branchout,xiao2025cld}, which is then fine-tuned by closed-loop reinforcement learning toward safety-critical interactions~\cite{black2024training}.
Third, \emph{attribution-aware generation} optimizes the adversarial policy for collisions whose responsibility can be assigned to the ADS under the CCD standard.

We validate the effectiveness and generalization of CARS across  multiple experiments. With regard to effectiveness, the majority of generated collisions are correctly attributed to the ADS under FSM, and the attribution remains consistent under the CC-JP and RSS cross-checks, indicating that the result is not specific to a single CCDM specification. With regard to generalization, the same generation policy transfers without retraining across three datasets spanning heterogeneous national traffic environments: nuScenes urban driving~\cite{caesar2020nuscenes}, AD4CHE Chinese highway  traffic~\cite{zhang2023ad4che}, and RounD German roundabout  interaction~\cite{krajewski2020round}.

In summary, these results show that adversarial simulation can construct responsibility-attributed test evidence, not only produce collisions. This work moves adversarial scenario generation from collision discovery to responsibility-attributed scenario generation, providing a methodology aligned with the regulatory governance framework for scalable ADS safety testing.

\section{Results}
\label{sec:2}

\subsection{Evaluation design}
\label{sec:experimental_setup}

We evaluate CARS by asking whether generated collisions are attributable to the ADS, physically feasible, severity-diverse, and transferable beyond the training setting.
The primary evaluation is conducted on the nuScenes urban-driving dataset~\cite{caesar2020nuscenes}.
Cross-domain transfer is then tested on highway traffic from AD4CHE~\cite{zhang2023ad4che} and roundabout interactions from RounD~\cite{krajewski2020round}, without retraining the adversary policy (dataset details in \sisec{si:datasets}; map in \sifig{fig:si_dataset_map}).
The primary nuScenes evaluation uses a Gaussian-mixture diffusion planner as the ADS under test; ADS-planner robustness experiments replace it with alternative planners.

Generated collisions are evaluated under three reference models with different modeling assumptions.
FSM serves as the primary attribution reference because it operationalizes the UN ECE\,R157 CCD standard with a fuzzy braking response that produces graduated decelerations rather than binary threshold decisions.
This continuous response also provides a natural severity scale for distinguishing Easy, Medium, and Hard encounters~\cite{unece2021r157}.
CC-JP and RSS are retained as cross-checks under different structural assumptions.
\Cref{tab:main_results} reports responsibility validity under the three reference models, severity diversity ($H_{\mathrm{crit}}$), positive braking-deficit exposure (BD$^+$\%, the fraction of scenarios with BD\,$>$\,0), and trajectory infeasibility (IP\%); definitions and per-axis kinematic checks are given in Methods and Supplementary Tables~\ref{si:tab:main_results}, \ref{si:tab:validity}, and \ref{tab:si_kinematics}.

\subsection{Adversary selection}
\label{sec:dt_eval}

Context-aware selection keeps the active adversary aligned with the evolving threat rather than fixing the attacker at the start of a scenario.
We evaluate the selector on validation target--adversary (\textit{tgt}--\textit{adv}) pairs and report the full classification and ranking statistics in \sitab{si:tab:dt_evaluation}.
Because several surrounding agents can be plausible threats in the same scene, ranking is the relevant test: the labeled \textit{adv} appears among the three highest-ranked agents in 97.8\% of validation \textit{tgt}--\textit{adv} pairs.
Candidate adversaries are rescored at every simulation step.
A temporal confirmation gate promotes a new \textit{adv} only after its score remains highest for several consecutive frames, preventing momentary score spikes from switching the policy away before the current \textit{adv} has shaped the interaction.
The selector is not intended to infer a unique adversarial intent from a single frame.
Its role is to keep the adversarial policy attached to the surrounding agents that currently forms the most relevant conflict with the ADS.

\begin{figure*}[t]
  \centering
  \includegraphics[width=\textwidth]{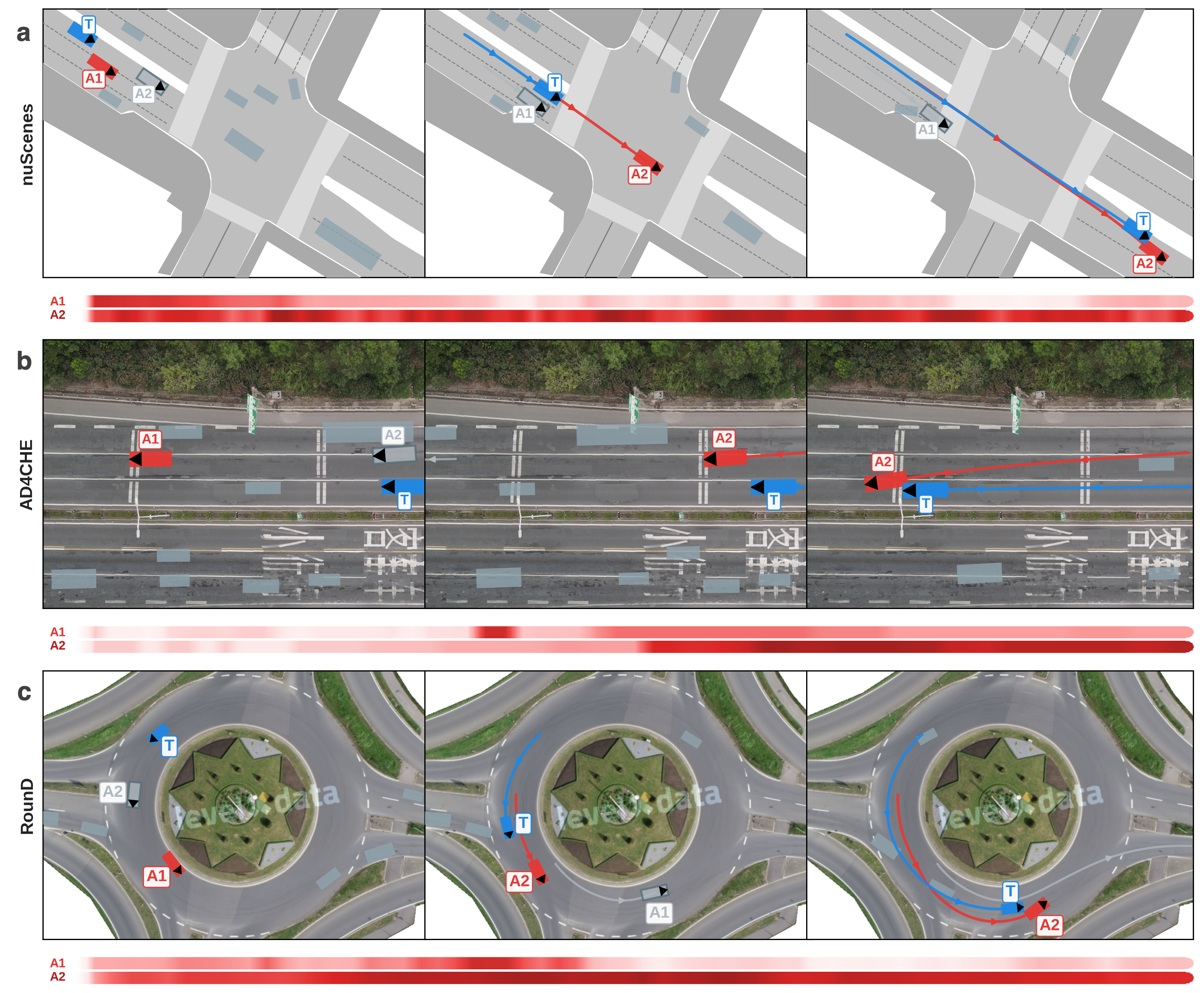}
  \caption{Context-aware \textit{adv} re-selection across datasets.
  \textbf{a}, nuScenes. \textbf{b}, AD4CHE. \textbf{c}, RounD.
  Each row shows three BEV snapshots from one rollout scenario at $t{=}0$, the \textit{adv} switch frame, and the collision frame, with a heatmap strip beneath giving the per-step adversarial probability $P_{\mathrm{adv}}(t)$ for exemplary candidates A1 (upper) and A2 (lower); darker red indicates higher probability.
  The \textit{tgt} is shown in \textcolor{blue}{blue}, the currently active \textit{adv} in \textcolor{red}{red}, and other agents in gray.}
  \label{fig:adv_switch}
\end{figure*}

\Cref{fig:adv_switch} shows how the selected adversary changes as the conflict develops.
In the nuScenes example, the initial \textit{adv} is replaced only after a closer agent sustains a higher adversarial score over the confirmation window; this newly selected \textit{adv} then closes on the \textit{tgt}.
The same selection rule identifies a lane-change threat in AD4CHE and a roundabout-entry yielding conflict in RounD, indicating that the mechanism tracks the evolving conflict rather than a fixed scene template.
Most rollouts retain the initially selected \textit{adv}, but 12--16\% require a switch.
These switches occur when the originally assigned \textit{adv} no longer dominates the interaction, showing why adversary selection must remain active during rollout.

\subsection{Responsibility attribution on nuScenes}
\label{sec:results}
\label{sec:validity}%

In the primary nuScenes evaluation, CARS retains 88.7\% attribution under the primary FSM reference (\Cref{tab:main_results}).
The auxiliary checks give 79.7\% attribution under CC-JP and 97.1\% under RSS, indicating that the attribution result is not specific to a single reference specification.
These auxiliary checks are interpreted as sensitivity tests rather than as a replacement for the primary FSM criterion.
As a stricter robustness subset, 73.8\% of the collisions are attributable under all three reference models.
Only 2.2\% are unpreventable under all three, so 97.8\% remain preventable by at least one reference model.
This distinction matters for validation: a generator that simply forces the adversary into unrecoverable geometry would raise unpreventable rates under all references, whereas CARS mostly produces encounters that remain avoidable under the primary reference or at least one auxiliary check.

\Cref{fig:results_analysis}a--c illustrates representative disagreements among the reference models.
Panels a--c show cases in which FSM, CC-JP, and RSS, respectively, would still collide while the other references recover.
These cases are not treated as a replacement for the primary FSM criterion; they show how auxiliary references expose sensitivity to model-specific avoidability assumptions.
If CARS were mainly producing geometrically impossible collisions, all three references would fail together.
Instead, the all-reference-unpreventable set remains small.
This pattern supports the interpretation that CARS produces challenging but still evaluable encounters, rather than collisions whose attribution is already lost by construction.

\subsection{Severity diversity and kinematic feasibility}
\label{sec:gmm_ablation}%

CARS-generated collisions span a broad range of FSM braking demand while remaining within the feasibility bounds used for IP.
Severity is summarized by the FSM Easy, Medium, and Hard criticality tiers, which describe the braking demand imposed on the FSM reference: Easy cases remain within a comfort-level response, whereas Hard cases approach the emergency-braking regime.
On nuScenes, generated scenarios populate all three tiers, yielding $H_{\mathrm{crit}}{=}0.798$ (\Cref{tab:main_results}; per-tier and per-method breakdowns in \sitab{si:tab:main_results} and \sitab{si:tab:criticality_distribution}).

\begin{figure*}[htbp]
  \centering
  \begin{subfigure}[t]{0.32\textwidth}
    \paneltitle{a}{FSM fails}
    \includegraphics[width=\linewidth]{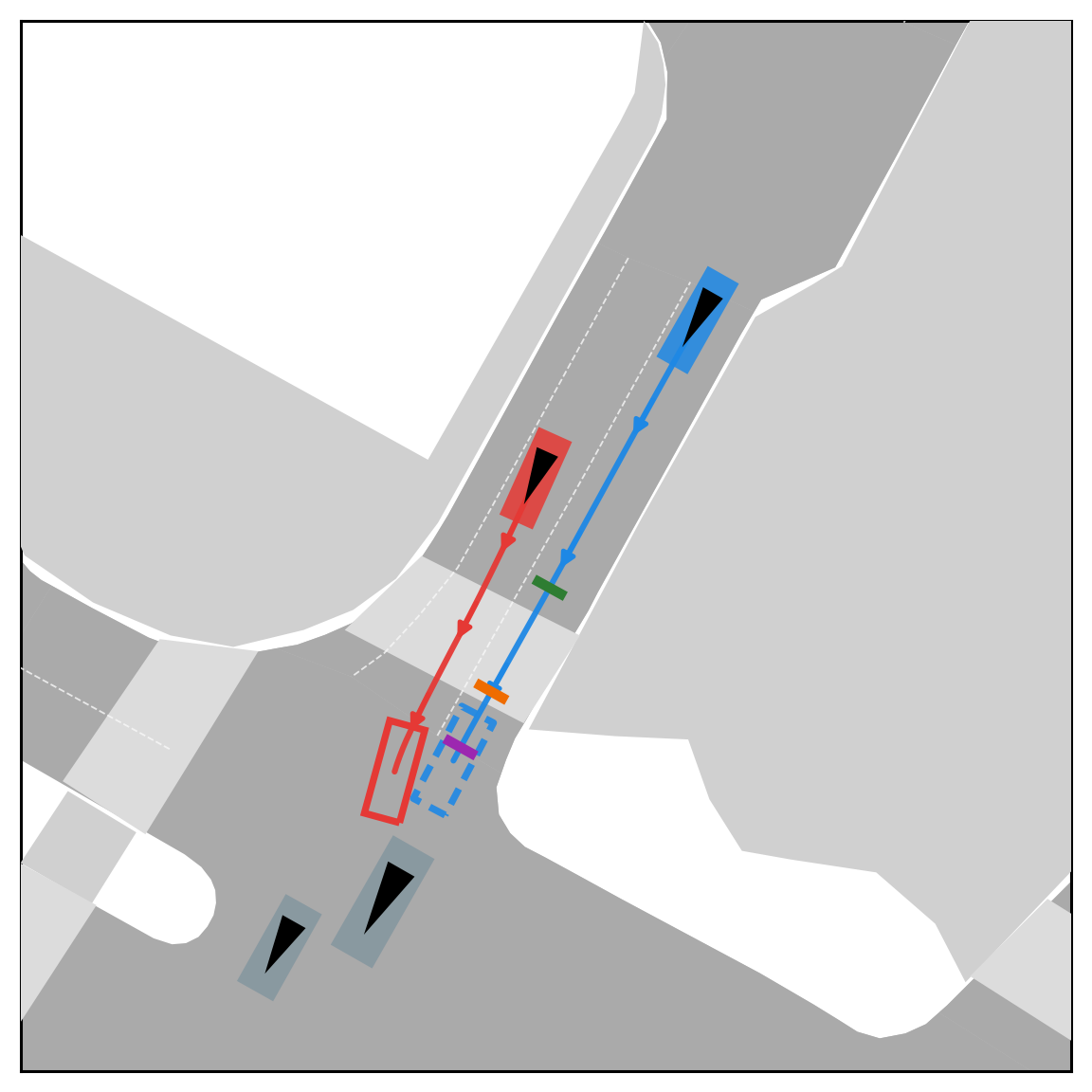}
  \end{subfigure}\hfill
  \begin{subfigure}[t]{0.32\textwidth}
    \paneltitle{b}{CC-JP fails}
    \includegraphics[width=\linewidth]{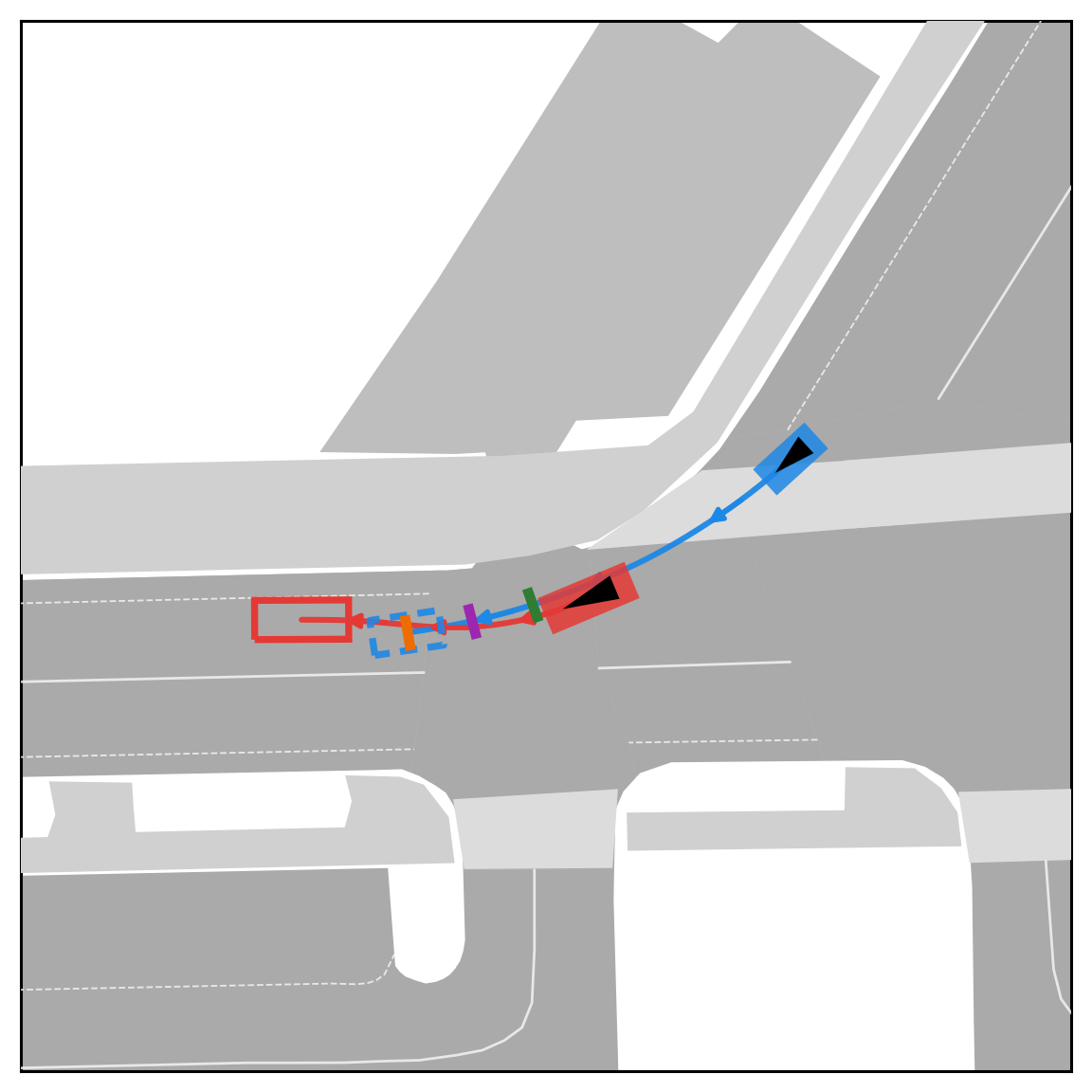}
  \end{subfigure}\hfill
  \begin{subfigure}[t]{0.32\textwidth}
    \paneltitle{c}{RSS fails}
    \includegraphics[width=\linewidth]{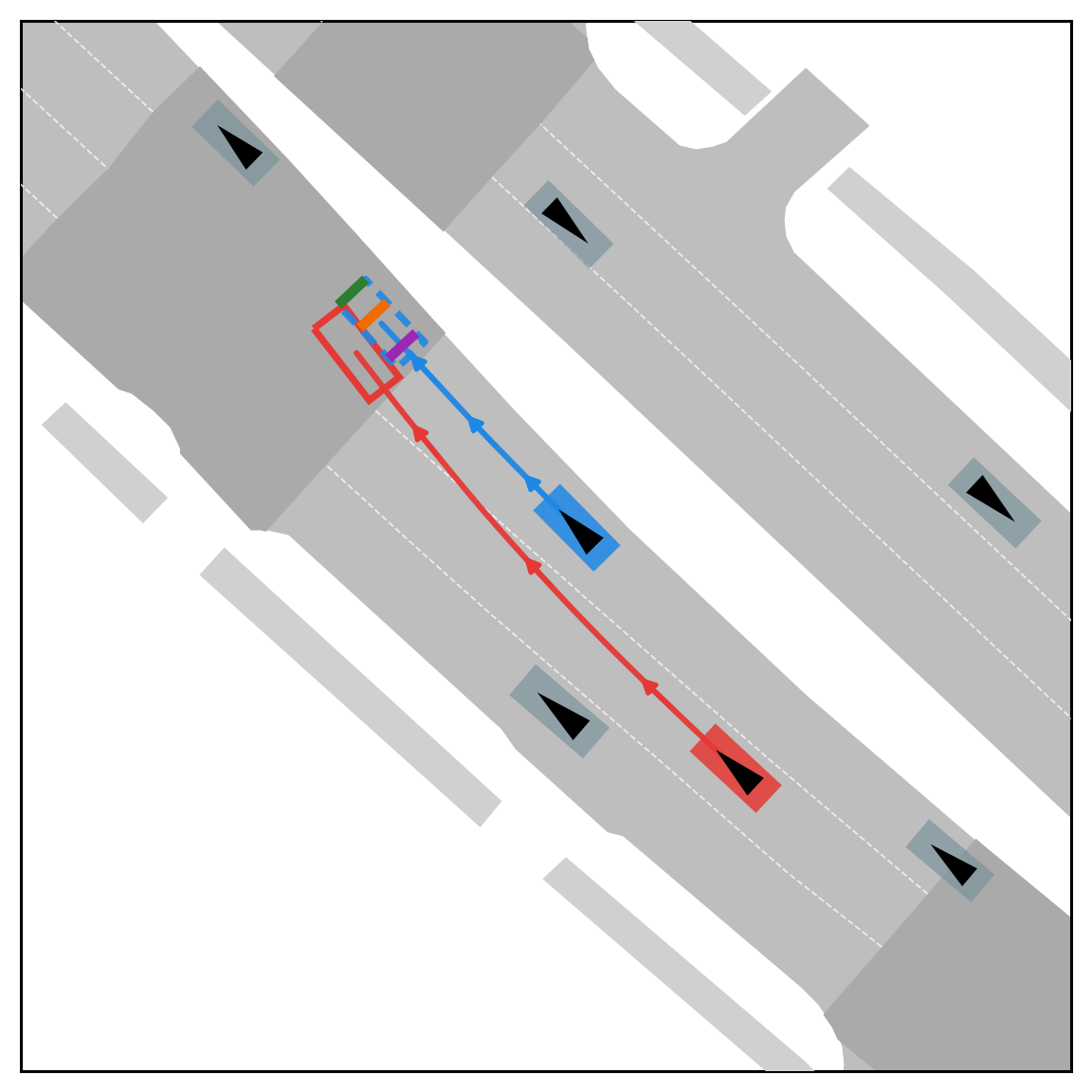}
  \end{subfigure}

  \vspace{4pt}

  \begin{minipage}{\textwidth}
    \centering
    \sffamily\footnotesize
    \setlength{\fboxsep}{5pt}%
    \fcolorbox{black!25}{white}{%
    \begin{tabular}{@{}l@{\hspace{14pt}}l@{\hspace{14pt}}l@{\hspace{18pt}}l@{\hspace{14pt}}l@{\hspace{14pt}}l@{}}
      \textcolor[HTML]{E53935}{\rule[-0.5pt]{1.0em}{0.65em}}\,Adversary &
      \textcolor[HTML]{1E88E5}{\rule[-0.5pt]{1.0em}{0.65em}}\,Target &
      \textcolor{gray}{\rule[-0.5pt]{1.0em}{0.65em}}\,Other agents &
      \textcolor[HTML]{9C27B0}{\rule[2pt]{1.0em}{1.6pt}}\,FSM &
      \textcolor[HTML]{EF6C00}{\rule[2pt]{1.0em}{1.6pt}}\,CC-JP &
      \textcolor[HTML]{2E7D32}{\rule[2pt]{1.0em}{1.6pt}}\,RSS \\
    \end{tabular}}
  \end{minipage}

  \vspace{6pt}

  \begin{subfigure}[t]{0.32\textwidth}
    \paneltitle{d}{Physical-margin distribution}
    \vspace*{3pt}
    \makebox[\linewidth][l]{\hspace*{-10pt}\includegraphics[width=\linewidth]{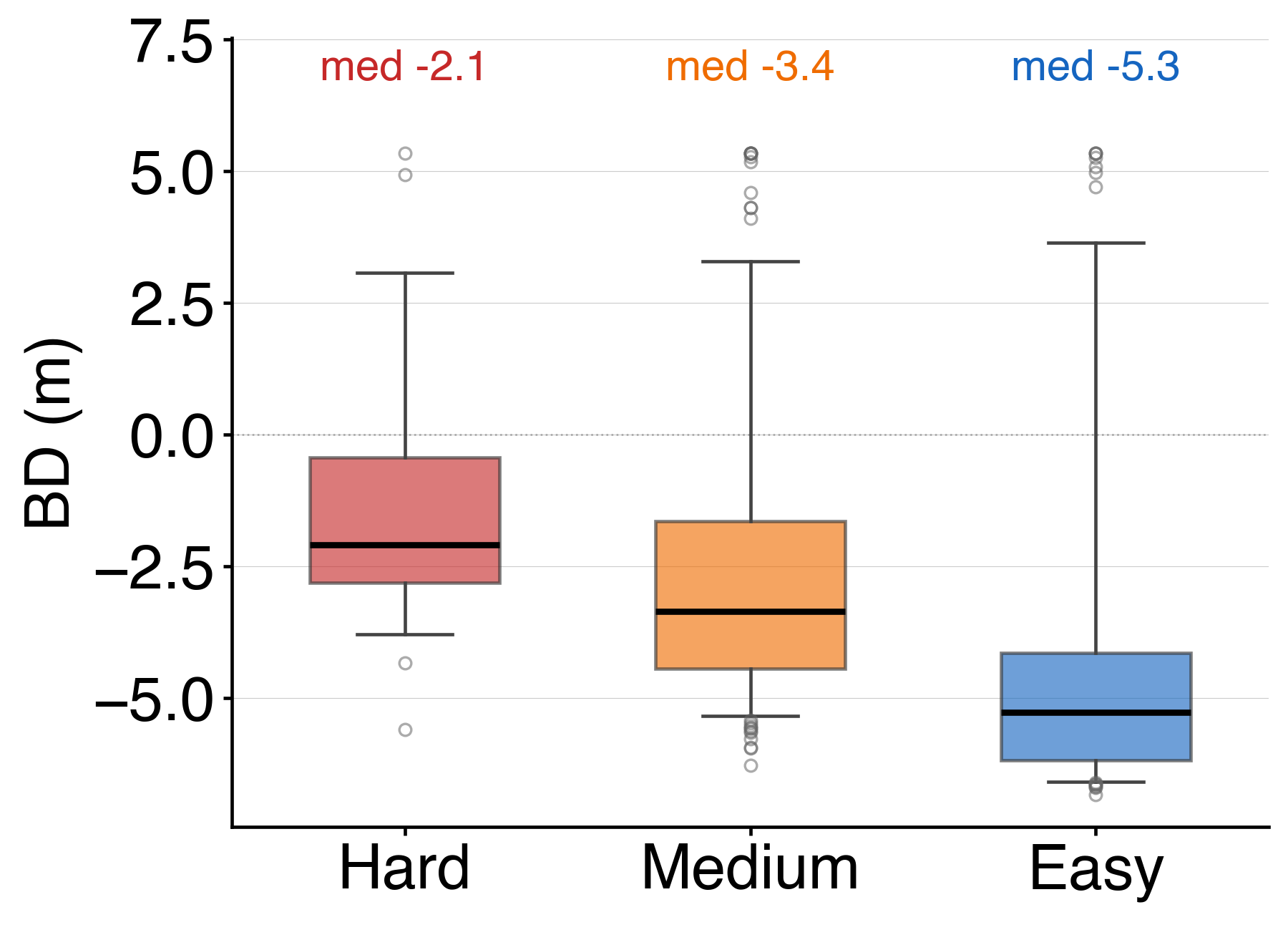}}
  \end{subfigure}\hfill
  \begin{subfigure}[t]{0.32\textwidth}
    \paneltitle{e}{\textit{Adv} speed at collision}
    \vspace*{3pt}
    \makebox[\linewidth][l]{\hspace*{-13pt}\includegraphics[width=\linewidth]{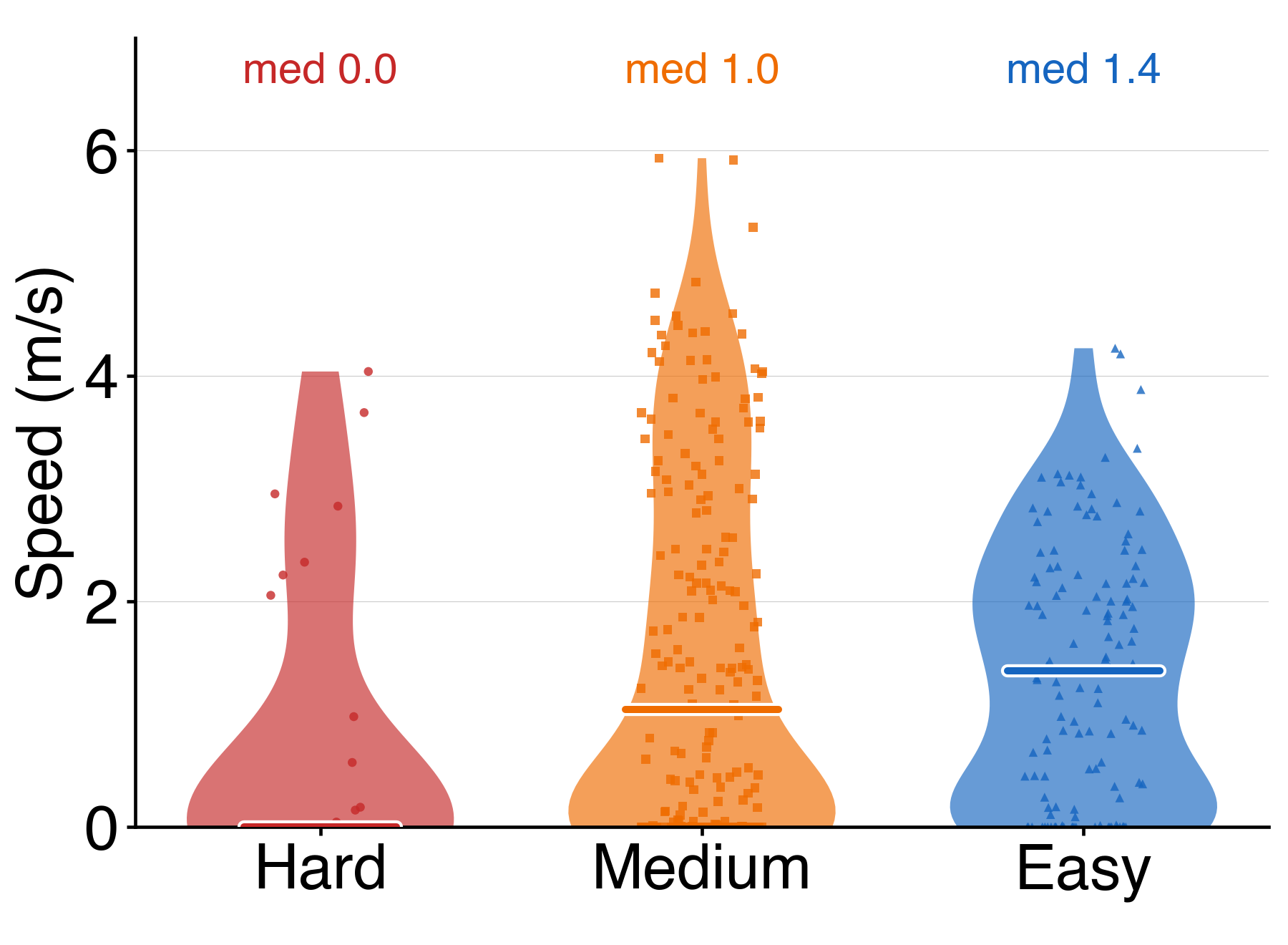}}
  \end{subfigure}\hfill
  \begin{subfigure}[t]{0.32\textwidth}
    \paneltitle{f}{Approach urgency}
    \vspace*{2pt}
    \makebox[\linewidth][l]{\hspace*{-6pt}\includegraphics[width=\linewidth]{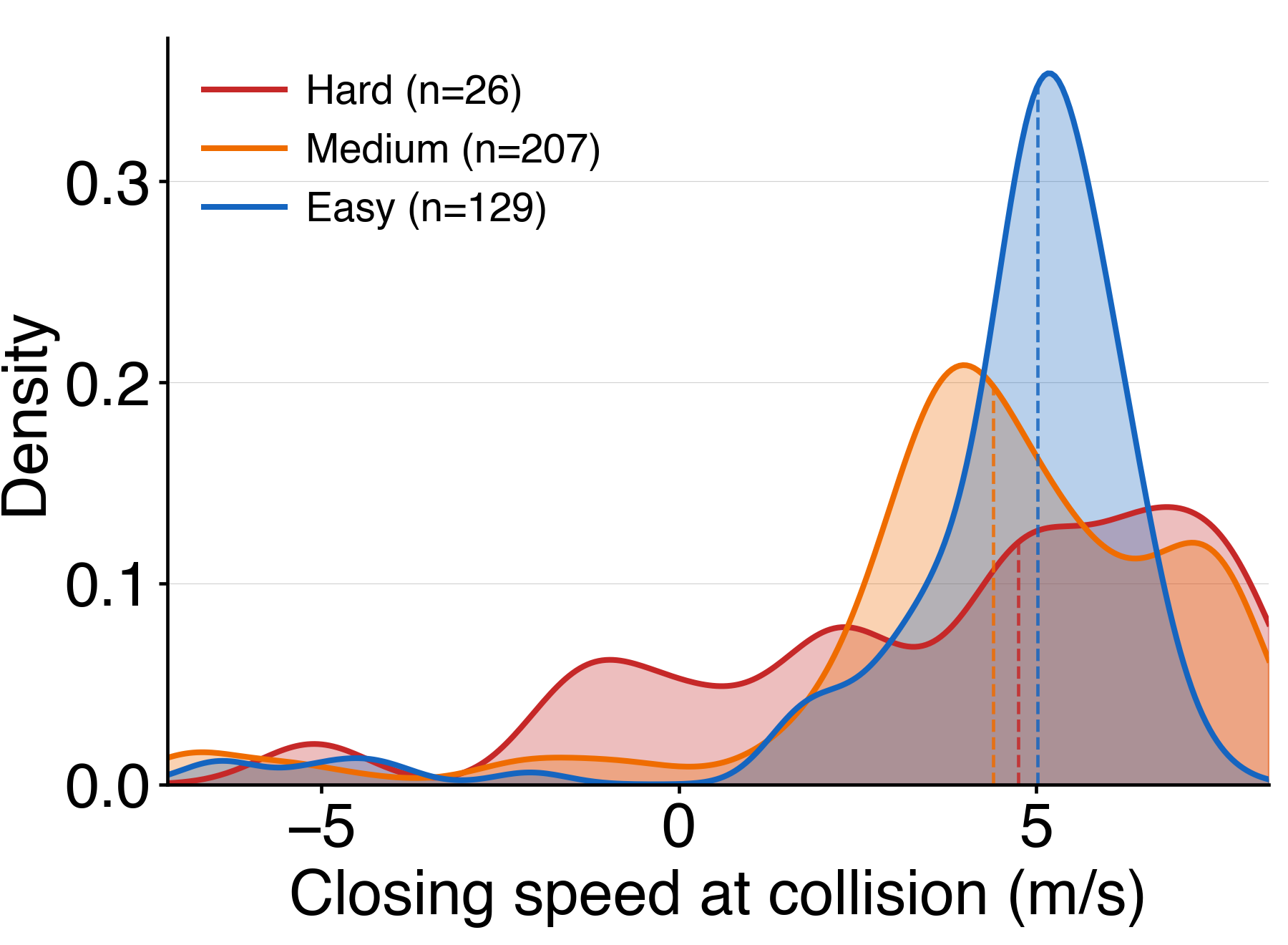}}
  \end{subfigure}

  \caption{Responsibility attribution and collision kinematics on CARS-generated nuScenes scenarios.
  \textbf{a--c}, Representative attribution disagreements in which FSM fails (\textbf{a}), CC-JP fails (\textbf{b}), and RSS fails (\textbf{c}). \textit{adv} trajectory in \textcolor{red}{red}; \textit{tgt} trajectory in \textcolor{blue}{blue}; the \textit{tgt}'s stop position under each reference model is shown as a front-edge line (\textcolor[HTML]{9C27B0}{purple}, FSM; \textcolor[HTML]{EF6C00}{orange}, CC-JP; \textcolor[HTML]{2E7D32}{green}, RSS), as keyed in the legend.
  \textbf{d}, Max braking deficit BD, box-and-whisker per FSM criticality tier (Easy / Medium / Hard); box spans the interquartile range, whiskers extend to the 5th–95th percentiles, points are outliers, and median is annotated. BD\,$>$\,0 indicates that required stopping distance exceeded the available gap.
  \textbf{e}, \textit{Adv} speed at collision, raincloud plot per FSM criticality tier (median annotated).
  \textbf{f}, Longitudinal closing speed at collision, kernel-density estimate per criticality tier (median annotated).}
  \label{fig:results_analysis}
\end{figure*}

\Cref{fig:results_analysis}d--f characterizes the collision kinematics behind this spread.
The BD distributions separate the FSM criticality tiers, confirming that the tier labels reflect differences in braking demand.
The \textit{adv} speed and longitudinal closing-speed panels show that collisions occur during sustained approaches, rather than from stationary or marginal contacts.
The Easy and Medium tiers remain populated, indicating that the generated set does not collapse into a single emergency-braking regime.
Additional nuScenes collision geometries are provided in \sisec{si:collision_gallery}.

The same scenarios remain physically plausible under the feasibility checks used for IP.
CARS has IP\% equal to 0.04\% in \Cref{tab:main_results}, and the percentile-level acceleration, jerk, and lateral-acceleration checks remain below the feasibility bounds (\sitab{tab:si_kinematics}).
Together, the severity and feasibility results show that CARS preserves a spread of attributable interaction regimes without relying on kinematic artifacts.

\subsection{Comparison with adversarial generators}
\label{sec:baseline_comparison}%

\begin{figure*}[htbp]
  \centering
  \begin{subfigure}[t]{\textwidth}
    \paneltitle{a}{Severity diversity}
    \includegraphics[width=\linewidth]{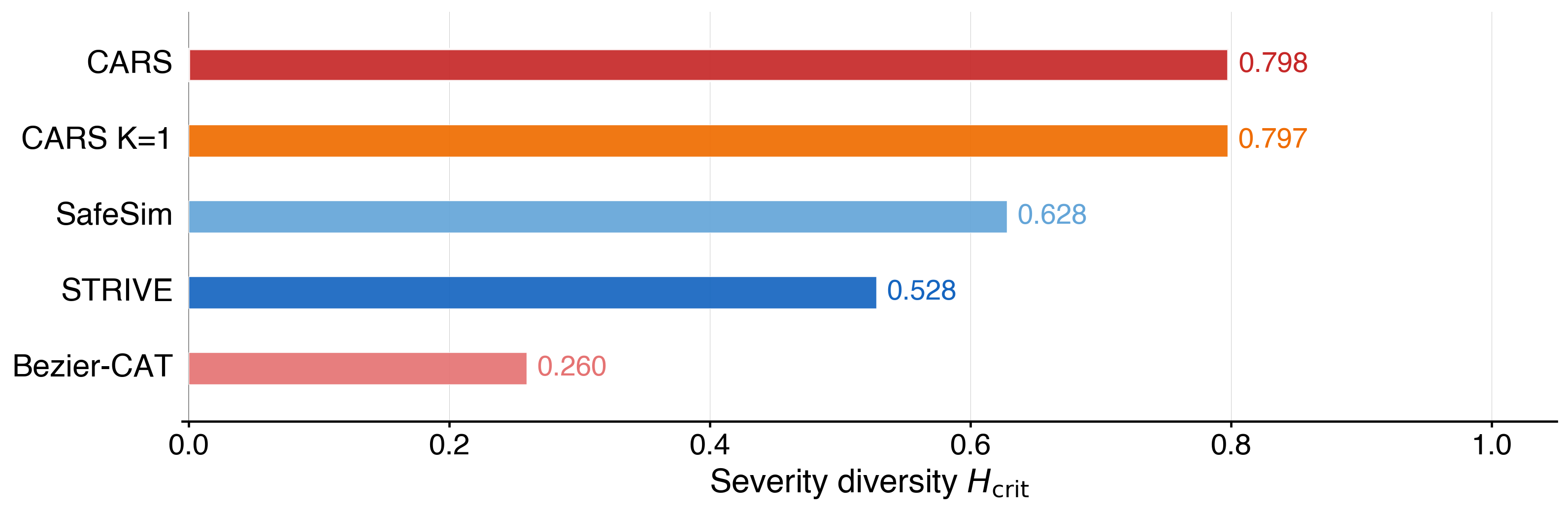}
  \end{subfigure}

  \vspace{6pt}

  \begin{subfigure}[t]{0.32\textwidth}
    \paneltitle{b}{Three-CCDM validity}
    \includegraphics[width=\linewidth]{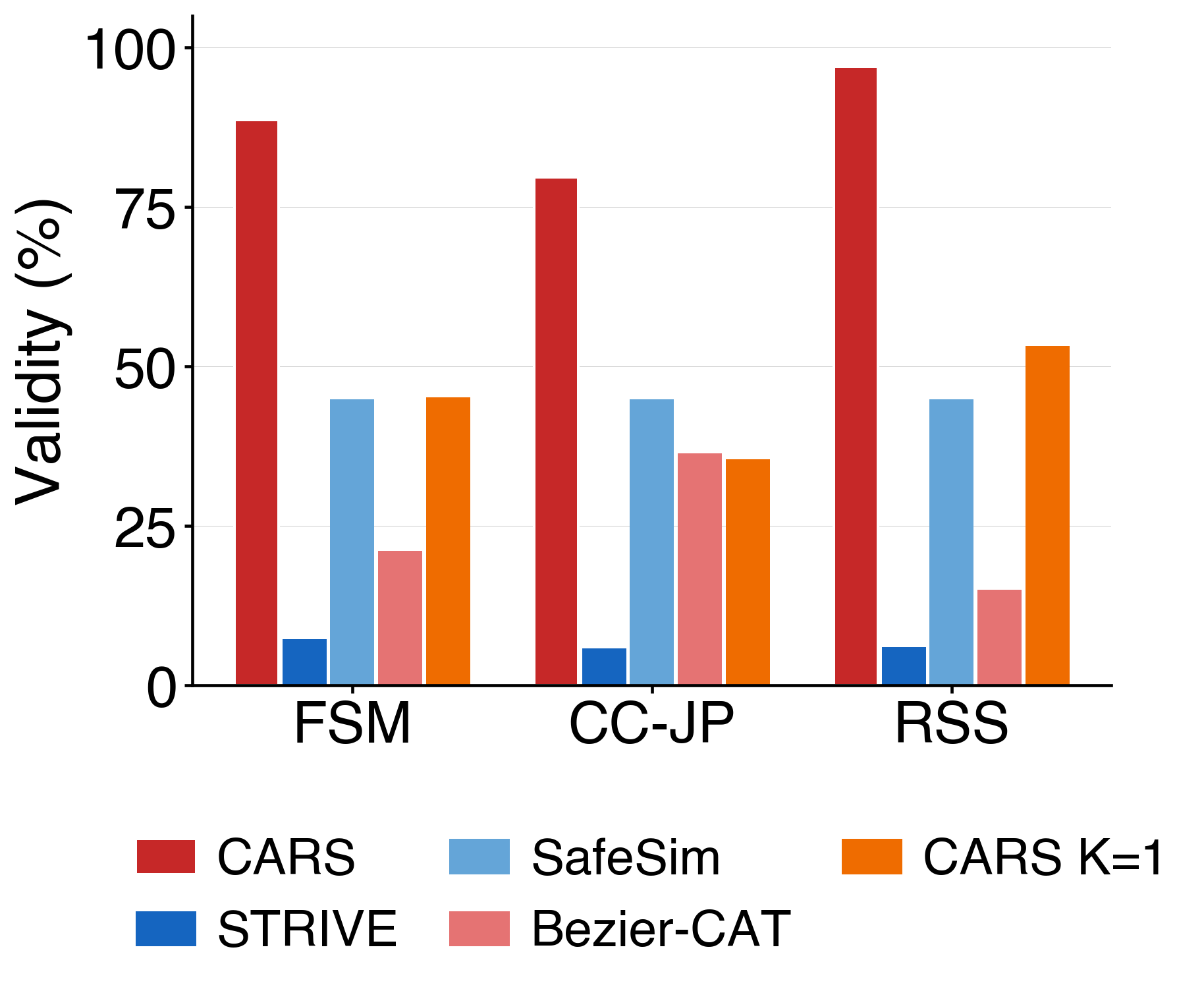}
  \end{subfigure}\hfill
  \begin{subfigure}[t]{0.32\textwidth}
    \paneltitle{c}{Validity-feasibility trade-off}
    \includegraphics[width=\linewidth]{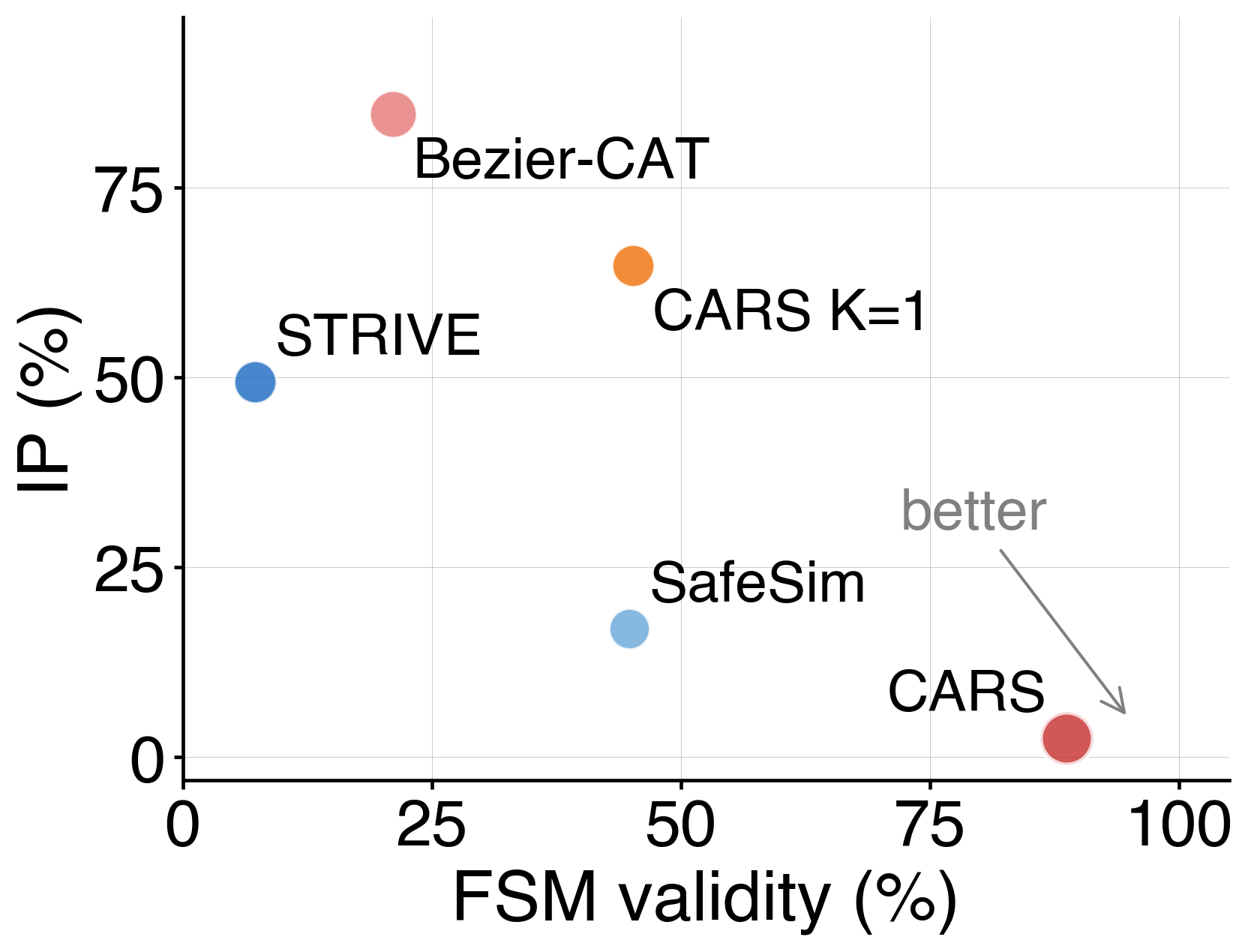}
  \end{subfigure}\hfill
  \begin{subfigure}[t]{0.32\textwidth}
    \paneltitle{d}{Severity tier coverage}
    \includegraphics[width=\linewidth]{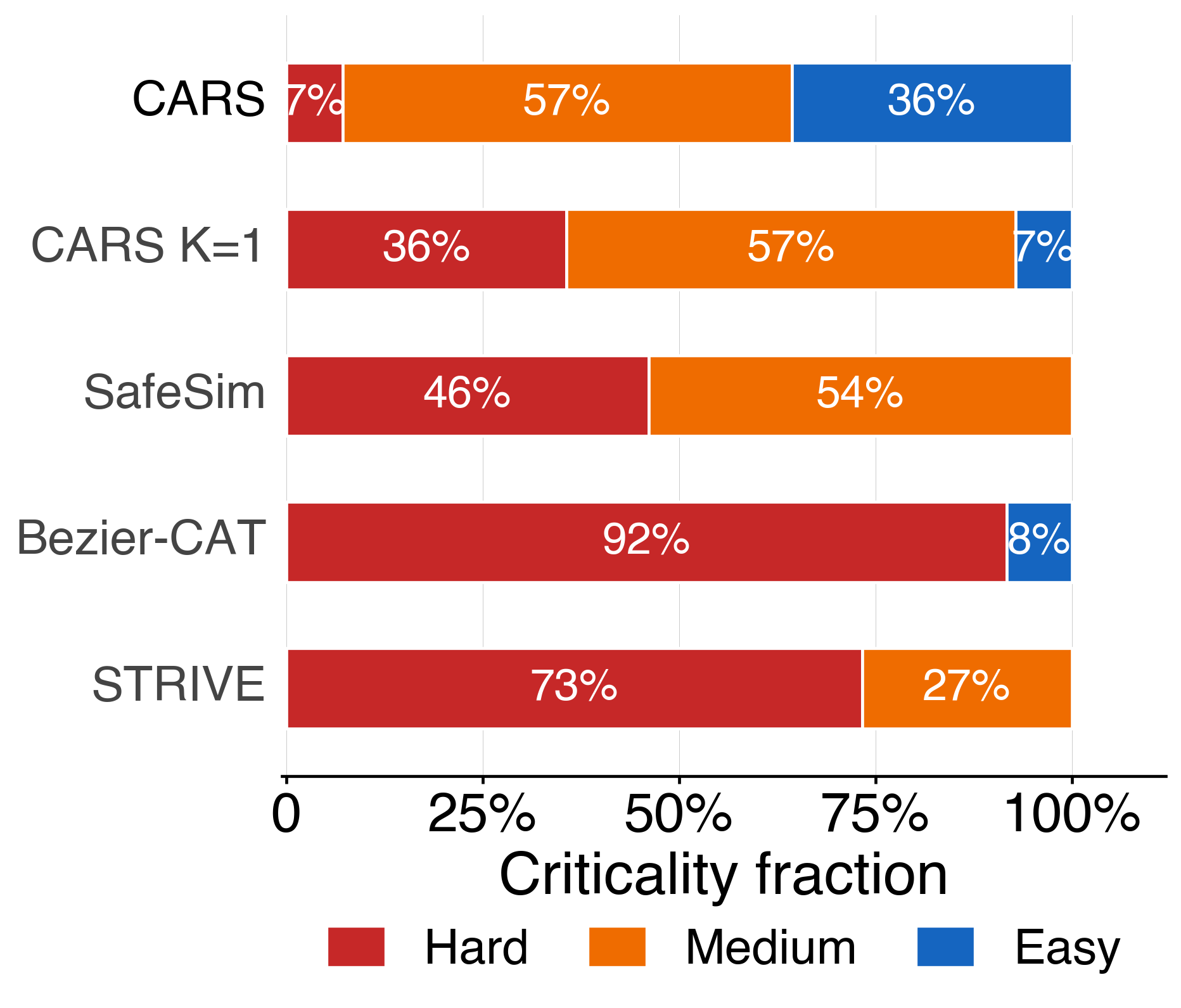}
  \end{subfigure}

  \caption{Baseline comparison on nuScenes.
  \textbf{a}, Severity-diversity entropy $H_\mathrm{crit}$ per method (raw point estimate on the FSM-preventable subset; same values as \Cref{tab:main_results}).
  \textbf{b}, Responsibility validity under FSM, CC-JP, and RSS per method, grouped bar.
  \textbf{c}, Validity-feasibility trade-off: FSM validity (\textit{x}) versus infeasibility percentage IP (\textit{y}); bubble area scales with the square root of the FSM-preventable scenario count. Lower-right is better.
  \textbf{d}, Per-method criticality tier coverage on the FSM-preventable subset, horizontal stacked bar.}
  \label{fig:baseline_comparison}
\end{figure*}

Under the same responsibility-attribution pipeline, collision-oriented generators do not automatically produce responsibility-attributable scenarios.
We compare CARS with STRIVE~\cite{rempe2022generating}, SafeSim~\cite{chang2024safe}, and a rule-based Bezier-CAT interception baseline inspired by CAT~\cite{zhang2023cat}, evaluating each method's generated collisions with the same FSM, CC-JP, and RSS attribution checks.
An internal CARS ($K{=}1$ \textit{adv}) ablation isolates the effect of replacing the adversarial policy's Gaussian-mixture output head with a one-component Gaussian output head.
\Cref{fig:baseline_comparison} visualizes the comparison, and \Cref{tab:main_results} reports the full metric matrix.

Under the same FSM criterion, CARS retains 88.7\% attribution, approximately twice the strongest external baseline, while also achieving the highest severity diversity and the lowest infeasibility rate among the compared methods (\Cref{fig:baseline_comparison}, \Cref{tab:main_results}).
The main metrics are stable across seven rollout seeds (\sifig{fig:si_seed_robustness}).
The baselines occupy different failure regimes.
STRIVE finds impacts that rarely remain attributable under any reference model.
SafeSim improves attribution relative to STRIVE, but still shows lower severity diversity and higher infeasibility than CARS.
The rule-based Bezier-CAT interception baseline has the highest BD$^+$ and IP rates, showing that geometric collision pressure can come at the cost of feasibility.
These failure modes differ, but they share the same limitation: collision occurrence alone does not imply usable validation evidence.

The internal CARS ($K{=}1$ \textit{adv}) ablation shows why raw severity diversity is not sufficient.
Replacing the adversarial policy's Gaussian-mixture output head with a one-component Gaussian output head leaves $H_{\mathrm{crit}}$ almost unchanged, but nearly halves FSM attribution and sharply increases infeasibility.
After requiring zero kinematic violation, only 25.0\% of FSM-preventable $K{=}1$ scenarios remain, whereas CARS retains 97.2\% with essentially unchanged severity diversity (\sitab{si:tab:feasible_criticality}).
Thus, the added trajectory capacity of the Gaussian-mixture adversary is expressed not as a broader severity histogram alone, but as severity diversity that survives attribution and feasibility checks.

\subsection{Generalization across planners and domains}
\label{sec:ablation}
\label{sec:cross_dataset}

The learned adversarial policy remains effective when the ADS-under-test planner is changed without retraining the adversarial policy.
On nuScenes, replacing the Gaussian-mixture diffusion planner used during training with a one-component diffusion planner leaves FSM attribution at 87.8\%, and replacing it with the architecturally distinct CTG planner~\cite{zhong2023guided} retains 86.0\% FSM attribution (\Cref{tab:main_results}).
Both planner-robustness tests preserve low infeasibility, with IP\% at 0.04\% and 0.02\%, respectively.
These results indicate that CARS does not depend on the particular ADS planner architecture used during training.

\begin{figure}[!htbp]
  \centering
  \begin{subfigure}[t]{\textwidth}
    \paneltitle{a}{Approach geometry across datasets}
    \includegraphics[width=\linewidth]{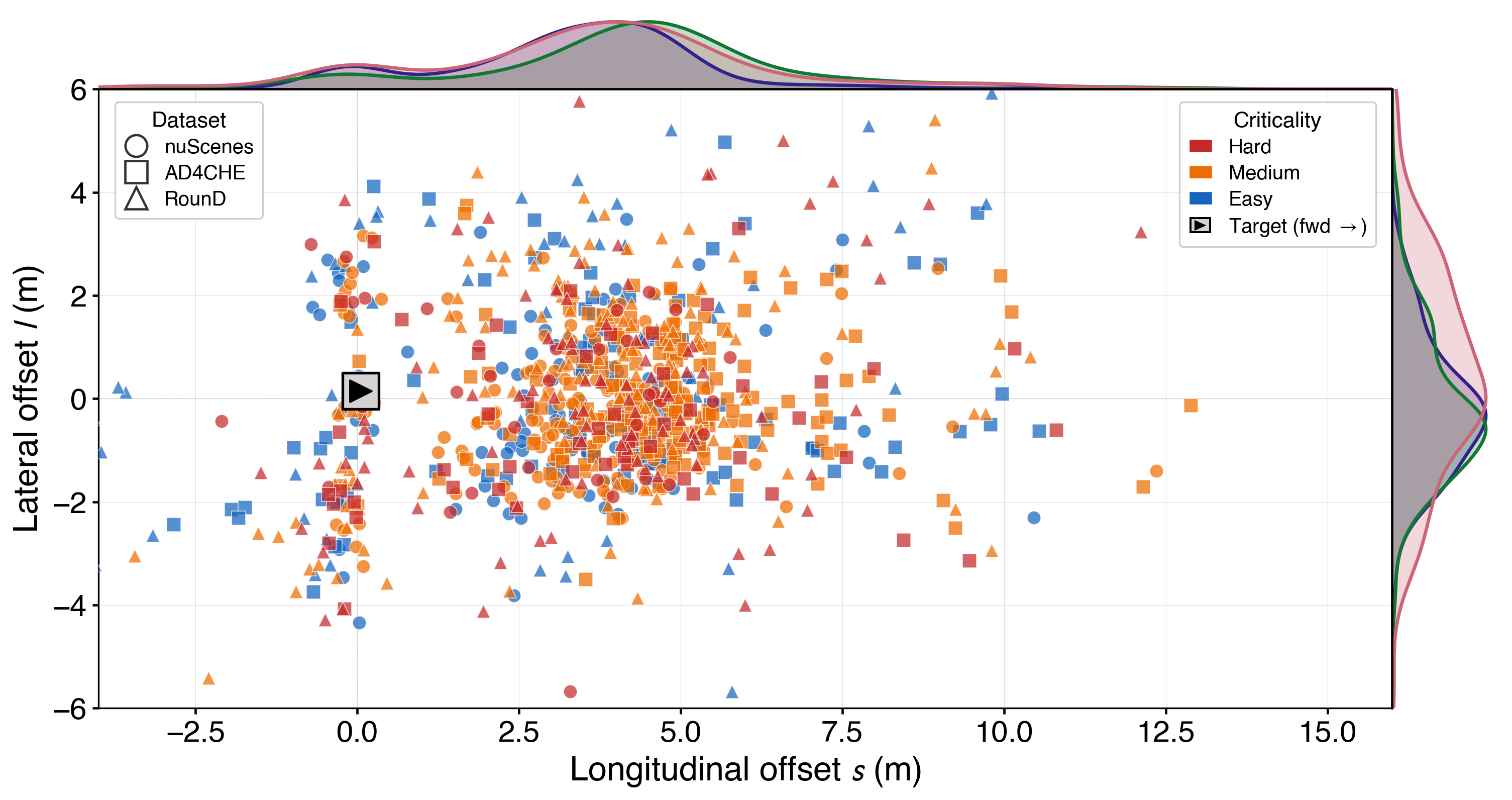}
  \end{subfigure}

  \vspace{6pt}

  \begin{subfigure}[t]{0.32\textwidth}
    \paneltitle{b}{CCDM validity (\%)}
    \includegraphics[width=\linewidth]{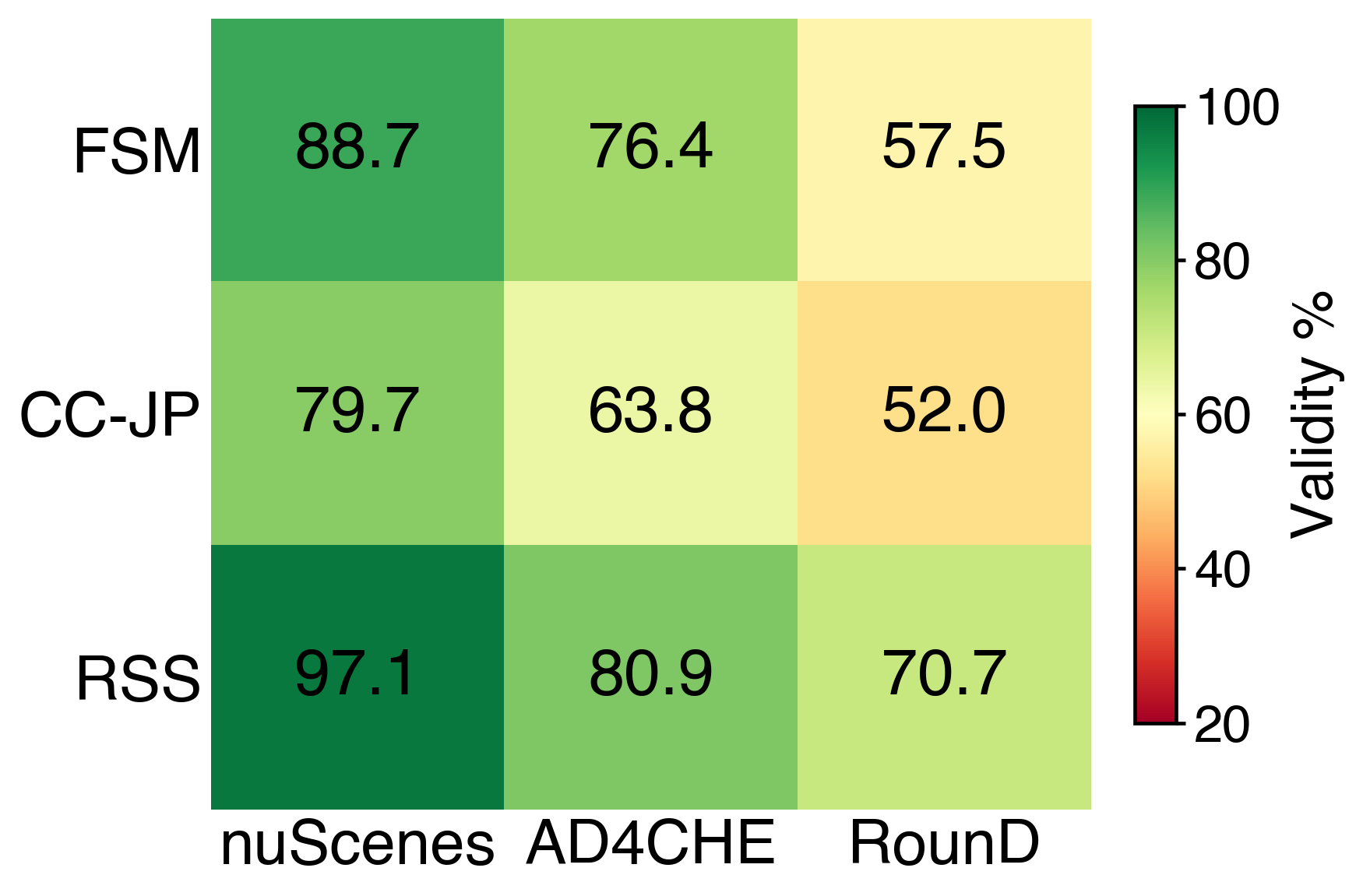}
  \end{subfigure}\hfill
  \begin{subfigure}[t]{0.32\textwidth}
    \paneltitle{c}{Kinematic urgency at impact}
    \includegraphics[width=\linewidth]{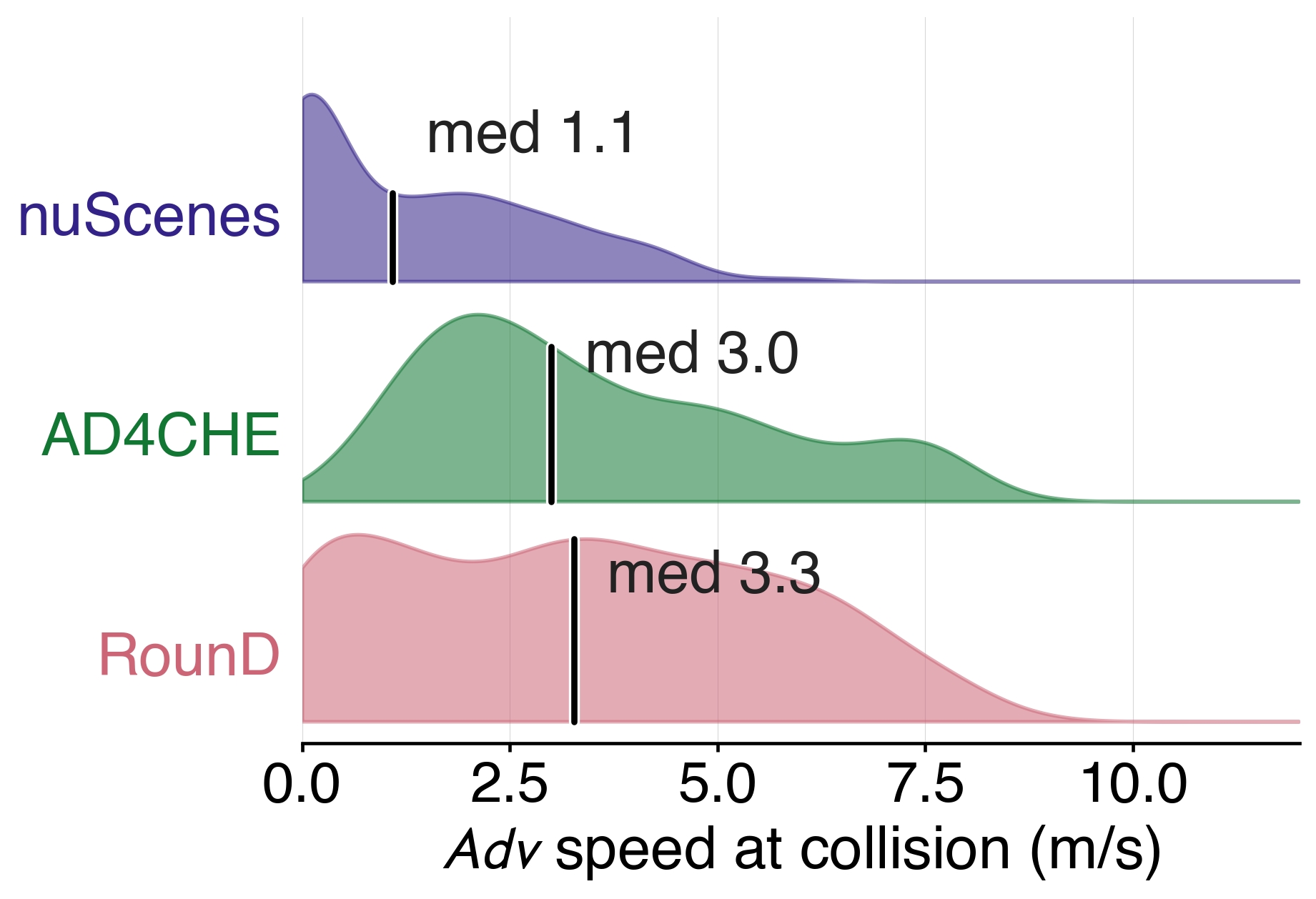}
  \end{subfigure}\hfill
  \begin{subfigure}[t]{0.32\textwidth}
    \paneltitle{d}{Severity tier shift}
    \includegraphics[width=\linewidth]{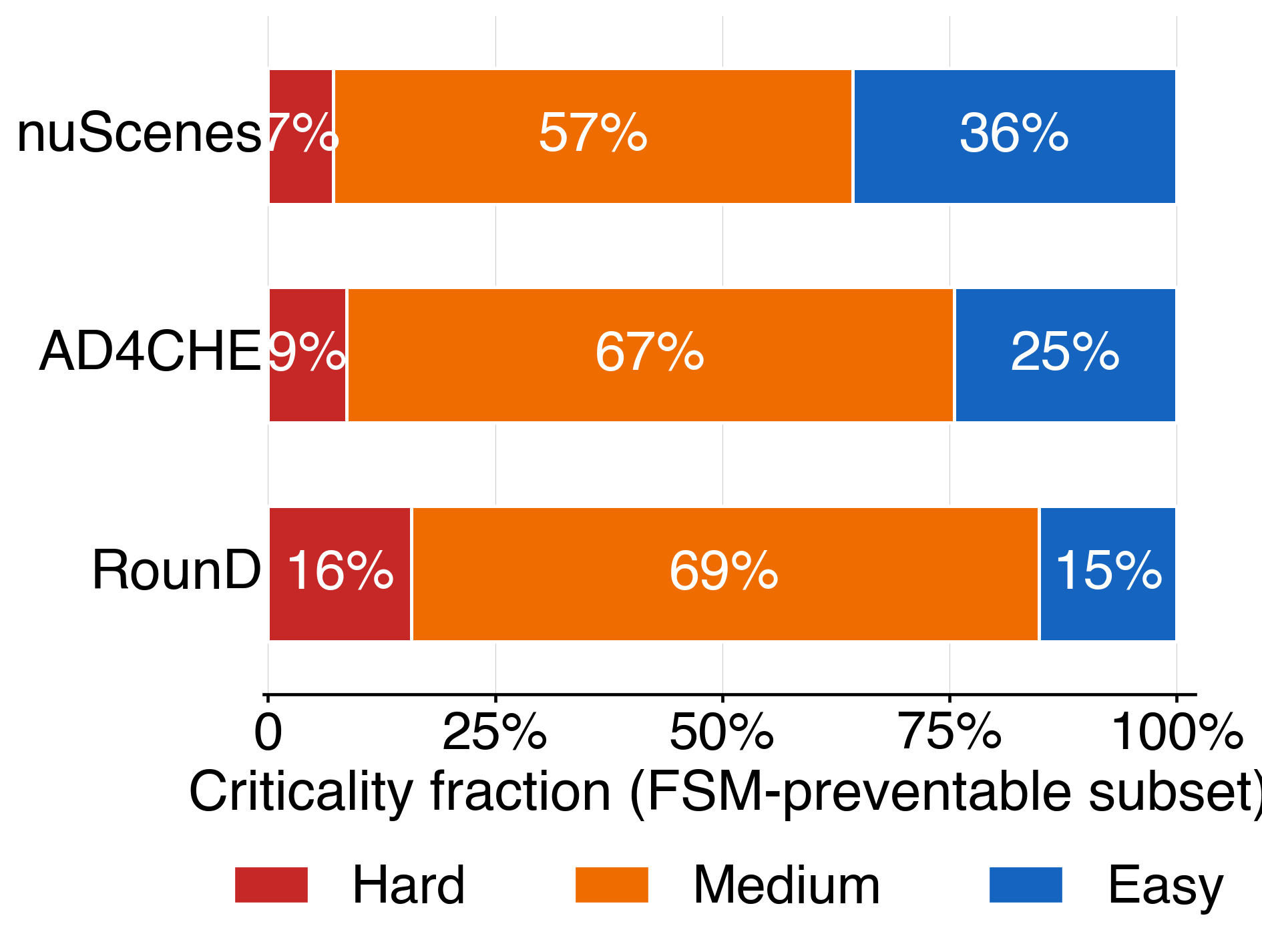}
  \end{subfigure}

  \caption{Cross-dataset generalization across nuScenes, AD4CHE, and RounD.
  \textbf{a}, Approach geometry overlaid across all three datasets, with shape encoding dataset ($\bigcirc$\,nuScenes; $\square$\,AD4CHE; $\triangle$\,RounD), color encoding FSM criticality tier (Hard/Medium/Easy), and 1D kernel-density marginals on top ($s$) and right ($l$).
  \textbf{b}, Responsibility validity under FSM, CC-JP, and RSS across the three datasets, color encoding validity \%.
  \textbf{c}, \textit{Adv} speed at collision, ridge plot per dataset.
  \textbf{d}, Severity tier shift across datasets, horizontal stacked bar on the FSM-preventable subset.}
  \label{fig:cross_dataset}
\end{figure}

We next deploy the same adversarial policy on AD4CHE and RounD without retraining.
AD4CHE tests highway interactions and RounD tests roundabout interactions, giving two traffic geometries outside the urban nuScenes training setting (\Cref{fig:cross_dataset}).
FSM attribution remains 76.4\% on AD4CHE and 57.5\% on RounD, while severity diversity remains close to the nuScenes result (\Cref{tab:main_results}).
Physical feasibility transfers cleanly to AD4CHE (IP\,=\,0.00\%) and remains low on RounD (IP\,=\,2.32\%), where curved approaches and laterally dominated interactions differ most from the training distribution.
Together, these results show that the adversarial policy retains measurable attribution and low infeasibility across ADS planners and traffic geometries, while the lower RounD attribution rate reflects the added difficulty of roundabout interactions.

\Cref{fig:cross_dataset} provides the geometric and metric context for the cross-domain tests.
Panel \textbf{a} overlays generated conflicts in a \textit{tgt}-centered relative-position frame, giving a qualitative domain map.
Attribution is quantified in panel \textbf{b} and \Cref{tab:main_results}: all three reference models avoid a smaller fraction of RounD collisions than nuScenes or AD4CHE collisions.
This drop is not specific to the FSM reference alone.
Panel \textbf{d} shows that RounD contains a larger proportion of Hard FSM-preventable cases, indicating that roundabout interactions shift the FSM-based attribution profile toward higher braking demand.
This pattern is consistent with the curved and laterally dominated geometry of RounD, where a longitudinal braking reference has fewer ways to resolve conflicts than in more aligned approach settings.
Together, \Cref{fig:cross_dataset} and \Cref{tab:main_results} show that cross-domain deployment preserves measurable attribution and low infeasibility, but the roundabout domain shifts the generated evidence toward harder and less easily attributable encounters.

\begin{table}[!htbp]
\centering
\caption{Unified evaluation of CARS across adversarial-method comparisons, ADS-planner robustness tests, and cross-dataset generalization. In the ADS-planner rows, the adversarial policy is fixed and only the ADS-under-test planner is replaced.}
\label{tab:main_results}
\renewcommand{\arraystretch}{1.30}
\footnotesize\sffamily
\setlength{\tabcolsep}{7pt}
\begin{tabular}{lcccccc}
\toprule
\textbf{Method / Setting}
 & \textbf{FSM\,\%}\,$\uparrow$
 & \textbf{CC-JP\,\%}\,$\uparrow$
 & \textbf{RSS\,\%}\,$\uparrow$
 & $\boldsymbol{H_{\mathrm{crit}}}$\,$\uparrow$
 & \textbf{BD$^+$\,\%}\,$\downarrow$
 & \textbf{IP\,\%}\,$\downarrow$ \\
\midrule
\multicolumn{7}{l}{\textit{Adversarial methods on nuScenes}} \\
STRIVE                            & \phantom{0}7.3 & \phantom{0}5.8 & \phantom{0}6.1 & 0.528 & 53.8 & 36.39 \\
SafeSim                           & 44.8 & 44.8 & 44.8 & 0.628 & 48.3 & 12.94 \\
Bezier-CAT                        & 21.1 & 36.4 & 15.0 & 0.260 & 84.1 & 73.08 \\
CARS ($K{=}1$ \textit{adv})       & 45.2 & 35.5 & 53.2 & 0.797 & 66.1 & 27.40 \\
\textbf{CARS}                     & \textbf{88.7} & \textbf{79.7} & \textbf{97.1} & \textbf{0.798} & \textbf{22.5} & \textbf{\phantom{0}0.04} \\
\addlinespace[4pt]
\multicolumn{7}{l}{\textit{ADS-planner robustness (fixed CARS \textit{adv})}} \\
One-component diffusion planner        & 87.8 & 80.0 & 96.7 & 0.834 & 21.7 & \phantom{0}0.04 \\
CTG planner~\cite{zhong2023guided} & 86.0 & 80.0 & 92.4 & 0.745 & 29.2 & \phantom{0}0.02 \\
\addlinespace[4pt]
\multicolumn{7}{l}{\textit{Cross-dataset generalization}} \\
CARS on AD4CHE                    & 76.4 & 63.8 & 80.9 & 0.751 & 36.6 & \phantom{0}0.00 \\
CARS on RounD                     & 57.5 & 52.0 & 70.7 & 0.759 & 50.9 & \phantom{0}2.32 \\
\bottomrule
\end{tabular}
\end{table}

\section{Discussion}
\label{sec:discussion}

CARS shows that adversarial simulation can generate collisions whose responsibility is attributable, not merely collisions whose occurrence can be counted.
Collision counts alone can mix ADS failures with infeasible adversarial motion or encounters that no reasonable reference response would avoid.
CARS addresses this by linking three generation-side choices: online \textit{adv} selection, multi-component trajectory generation, and closed-loop reward shaping against a CCD reference.
The resulting scenarios remain largely attributable under the primary FSM reference, retain severity diversity after feasibility checks, and remain measurable when the ADS planner or traffic domain changes.

The main implication is that adversarial scenario generation should be evaluated as evidence construction.
For ADS validation, a generated collision is useful only if it supports a claim about the ADS under test.
That claim requires an independent reference response.
If the reference also collides, the scenario is a weak failure case for the ADS.
If the reference avoids the collision, the scenario becomes interpretable as an attributable failure.
This changes the objective from inducing impacts to constructing collisions that survive attribution, severity, and feasibility checks.

This framing also clarifies the role of reference models.
UN\,R157 defines ADS performance relative to a CCD, and FSM provides one operational longitudinal form of that standard.
CC-JP and RSS expose how the attribution decision changes under different reference assumptions.
Their disagreements are not a nuisance to hide.
They identify boundary cases where responsibility depends on the chosen reference model.
Reporting this structure is more informative than collapsing all generated collisions into one failure rate.

The present study has several limits.
The reference models are longitudinal, so the responsibility estimate is strongest for longitudinal conflicts.
Roundabout and laterally dominated interactions can shift the FSM-based criticality profile because a braking-only reference has fewer ways to resolve the encounter.
CARS also selects one active \textit{adv} at a time, leaving coordinated multi-agent attacks outside the present evaluation.
All evaluated agents are vehicles; vulnerable road users would require class-specific dynamics and responsibility assumptions.

Future work should extend responsibility-attributed generation beyond these boundaries.
Multi-agent adversarial policies could generate coordinated threats.
Reference models for lateral, crossing, and multi-agent conflicts would broaden the set of attributable scenarios.
Vehicle--pedestrian and vehicle--cyclist interactions will require different feasibility envelopes and responsibility rules.
As ADS regulation moves beyond longitudinal Automated Lane Keeping Systems (ALKS) settings, adversarial generators should be coupled to explicit reference standards so that generated scenarios remain useful as validation evidence.

\section{Methods}
\label{sec:method}

CARS is a closed-loop adversarial scenario generation framework with online \textit{adv} selection, diffusion-based adversarial generation, and CCD-based attribution verification.
First, a context-aware selector identifies the surrounding agent most likely to form a responsibility-relevant conflict with the ADS.
Second, CARS trains a Gaussian-mixture diffusion model to generate \textit{adv} action sequences conditioned on traffic context.
The model is then fine-tuned with reinforcement learning in closed-loop simulation to create sustained approach conflicts rather than isolated contacts.
Finally, attribution is verified with a CCD reference by holding the generated \textit{adv} trajectory fixed and replacing the \textit{tgt} response.
This section describes the main components; detailed adversary-selection features, diffusion losses, reinforcement learning fine-tuning, and the identifiability argument are provided in the Supplementary Information.

\subsection{Scenario formulation}
\label{sec:problem_formulation}

We consider a multi-agent traffic scene with agents $\mathcal{A}=\{1,\dots,N\}$.
Let $i_{\textit{tgt}}$ denote the agent controlled by the ADS under test; we refer to this agent as \textit{tgt}.
Each agent $i$ has state $s_t^i=[x_t^i,y_t^i,v_t^i,\theta_t^i]$, containing position, speed, and heading, and action $a_t^i=[\dot v_t^i,\dot\theta_t^i]$, containing longitudinal acceleration and yaw rate.
State transitions follow unicycle dynamics,
\begin{equation}
  s_{t+1}^{i}=f(s_t^i,a_t^i).
  \label{eq:dynamics}
\end{equation}
At each simulation step, agents are partitioned into the \textit{tgt}, the \textit{adv}, and background agents $\mathcal{A}_{\mathrm{bg}}(t)$.
The policies for \textit{tgt} and $\mathcal{A}_{\mathrm{bg}}(t)$ remain fixed during adversarial generation; only the policy controlling the active \textit{adv} is optimized.

Let $\mathcal{C}(\xi)\in\{0,1\}$ indicate whether rollout $\xi$ contains a collision.
For a generated \textit{adv} trajectory $\mathcal{T}_{\textit{adv}}$, let $\xi_{\mathrm{ADS}}(\mathcal{T}_{\textit{adv}})$ denote the original rollout with the ADS under test controlling \textit{tgt}, and let $\xi_{\mathrm{CCD}}(\mathcal{T}_{\textit{adv}})$ denote the counterfactual rollout in which the \textit{tgt} response is replaced by a CCD reference while $\mathcal{T}_{\textit{adv}}$ is fixed.
The collision is classified as ADS-attributable when
\begin{equation}
\mathbb{I}_{\mathrm{attr}}(\mathcal{T}_{\textit{adv}})
=
\mathbb{I}\!\left[
\mathcal{C}\!\left(\xi_{\mathrm{ADS}}(\mathcal{T}_{\textit{adv}})\right)=1
\;\land\;
\mathcal{C}\!\left(\xi_{\mathrm{CCD}}(\mathcal{T}_{\textit{adv}})\right)=0
\right].
\label{eq:ads_attribution}
\end{equation}

\subsection{Context-aware adversary selection}
\label{sec:decision_tree}

The active \textit{adv} is selected online rather than fixed at simulation start.
For each surrounding agent $i$, CARS computes a 13-dimensional feature vector $\boldsymbol{\phi}_{i,t}$ describing its relative geometry and kinematics with respect to \textit{tgt}, including distance, relative speed, heading alignment, longitudinal and lateral offsets, closing rates, bumper-to-bumper gap, time-to-collision, bearing angle, and short-horizon approach trends (definitions in \sisec{si:adv_features}).
A histogram-based gradient boosting classifier estimates the adversarial likelihood $P_{\mathrm{adv}}(i\mid\boldsymbol{\phi}_{i,t})$, and the highest-ranked candidate at step $t$ is
\begin{equation}
  \hat i_t = \arg\max_{i\in\mathcal{A}\setminus\{i_{\textit{tgt}}\}}
  P_{\mathrm{adv}}(i\mid\boldsymbol{\phi}_{i,t}).
  \label{eq:p1_selection}
\end{equation}
Candidate scores are recomputed at every control step so that the active \textit{adv} can follow the evolving threat.
To avoid oscillation caused by transient geometric changes, a new candidate is promoted only after $K_{\mathrm{conf}}$ consecutive steps of agreement:
\begin{equation}
 i_{t+1}^*=\begin{cases}
 \hat i_t, & \text{if } \hat i_{t-j}=\hat i_t\ \forall j\in\{0,\dots,K_{\mathrm{conf}}-1\},\\
 i_t^*, & \text{otherwise},
 \end{cases}
 \label{eq:confirm_gate}
\end{equation}
where $i_t^*$ is the active \textit{adv} index after temporal confirmation, with $i_0^*=\hat i_0$ at rollout start.

\subsection{Gaussian-mixture adversarial generation}
\label{sec:trajectory_generation}

Diffusion models (DMs) are used across ADS stress testing, from controllable perception attacks to traffic-scenario generation~\cite{zheng2025cadiffusion, xu2025diffscene}.
CARS first trains a DM to generate \textit{adv} action sequences conditioned on map, \textit{tgt} history, and neighbor history~\cite{ho2020denoising, janner2022planningdiffusionflexiblebehavior, jiang2023motiondiffuser, xu2025diffscene}.
Given context $\mathbf{c}$, the generator represents the future action sequence $\tau_a=\{a_{t+1},\dots,a_{t+T}\}$; the forward diffusion process is given in \sisec{si:forward_diffusion}.
Instead of predicting a single Gaussian denoising distribution, CARS uses a Gaussian-mixture diffusion model (GMDM) to represent a richer conditional action-sequence distribution, following recent mixture-parameterized diffusion and planning models for trajectory generation~\cite{chen2025gaussian, kim2025branchout}.
For noisy action sequence $\tau_a^r$ at diffusion step $r$, the denoising model predicts
\begin{equation}
  p_\theta(\tau_a^0\mid\tau_a^r,\mathbf{c},r)=
  \prod_{h=1}^{T}\left[\sum_{k=1}^{K}\pi_{h,k}
  \mathcal{N}(\tau_{a,h}^0;\mu_{h,k},\Sigma_{h,k})\right],
  \label{eq:gmm_denoising}
\end{equation}
where $h$ indexes the planning horizon and $\pi_{h,k}$, $\mu_{h,k}$, and $\Sigma_{h,k}$ are the mixture weight, mean, and covariance of Gaussian component $k$.
Training minimizes a negative log-likelihood objective with mixture-usage and component-separation regularizers,
\begin{equation}
\mathcal{L}_{\mathrm{total}}=\mathcal{L}_{\mathrm{NLL}}-\lambda_H\mathcal{H}_\pi+
\lambda_{\mathrm{repel}}\mathcal{L}_{\mathrm{repel}},
\label{eq:total_loss}
\end{equation}
with detailed definitions in \sisec{si:loss_components}.

The pretrained GMDM is then fine-tuned by reinforcement learning in closed-loop simulation, shifting its output distribution toward action sequences that create collision pressure.
This refinement is implemented with proximal policy optimization (PPO)~\cite{schulman2017proximal, Schulman2015HighDimensionalCC}.
The reward combines longitudinal progress toward \textit{tgt}, lateral proximity, lateral closing, a terminal collision bonus, and a time penalty:
\begin{equation}
R_{\mathrm{total}}=w_{\mathrm{prog}}R_{\mathrm{prog}}+w_\ell R_\ell+w_{\ell c}R_{\ell c}+R_{\mathrm{hit}}-\lambda_{\mathrm{time}}.
\label{eq:reward_total}
\end{equation}
These dense terms encourage \textit{adv} to enter and remain within the \textit{tgt}-relative conflict geometry before impact, so that generated collisions arise from sustained approach dynamics.
The reinforcement-learning objective and reward components are given in \sisec{si:ppo}.

\subsection{Responsibility attribution}
\label{sec:validity_verification}

Attribution is verified by holding the generated \textit{adv} trajectory fixed and replacing the \textit{tgt} response with an independent reference model.
We use FSM as the primary attribution reference for two reasons.
First, it gives an operational form of the UN\,R157 CCD standard for longitudinal ALKS conflicts.
Second, its fuzzy surrogate safety measures produce graduated braking responses, which are more consistent with graded driver braking behavior in ADS testing than a binary safe/unsafe switch~\cite{mattas2020fuzzy, mattas2022fuzzy, unece2021r157}.
CC-JP and RSS are used as auxiliary checks with different structural assumptions: CC-JP provides an alternative CCD reference from the same regulatory context, while RSS provides a formal safety envelope.
Reporting all three references tests whether attribution results depend on a single reference specification.

For FSM attribution, \textit{tgt} follows its original recorded path while its speed along that path is controlled by the FSM braking response (\sisec{si:roller_coaster}).
The generated \textit{adv} trajectory remains fixed.
FSM combines a Proactive Fuzzy Surrogate Safety Metric (PFS), which assesses whether \textit{tgt} can stop if \textit{adv} brakes hard, and a Critical Fuzzy Surrogate Safety Metric (CFS), which assesses imminent collision risk under current accelerations.
The saturated membership functions and safe/unsafe distance definitions for PFS and CFS are given in \sisec{si:fsm_full_spec}.
FSM then uses the Sugeno reaction rule,
\begin{equation}
b_{\textit{tgt}}=
\begin{cases}
\mathrm{CFS}\,(b_{\textit{tgt},\max}-b_{\textit{tgt},\mathrm{comf}})+b_{\textit{tgt},\mathrm{comf}}, & \mathrm{CFS}>0,\\[2pt]
\mathrm{PFS}\,b_{\textit{tgt},\mathrm{comf}}, & \mathrm{CFS}=0,
\end{cases}
\label{eq:fsm_decel}
\end{equation}
where $b_{\textit{tgt}}$ is the commanded positive deceleration magnitude, and $b_{\textit{tgt},\mathrm{comf}}$ and $b_{\textit{tgt},\max}$ denote the comfort and maximum deceleration limits.
The command is then followed by the regulatory reaction delay and jerk-limited ramp.
The same PFS/CFS values define the FSM criticality tiers used in the Results: Easy (PFS$\leq0.85$), Medium (PFS$>0.85$ and CFS$<0.9$), and Hard (CFS$\geq0.9$).
A generated collision is attributed to the ADS under a reference model when the reference-controlled \textit{tgt} avoids the collision in the counterfactual rollout.

FSM attribution should therefore be read as a longitudinal attribution claim.
It assumes that the generated \textit{adv} trajectory respects the relevant kinematic envelope, that the path-constrained counterfactual rollout is valid for the encounter, and that the conflict belongs to the longitudinal class addressed by UN\,R157.
Under these conditions, an FSM-attributed collision provides evidence that the ADS failed in an encounter that a CCD-style longitudinal reference could avoid; the formal lower-bound statement is provided in Supplementary Section~\ref{si:proof_thm1}.

\subsection{Implementation details}
\label{sec:implementation}

The GMDM backbone uses a ResNet-18 map encoder and a temporal convolutional denoising network conditioned on map context, \textit{tgt} history, and neighbor history.
The Gaussian-mixture head uses $K{=}8$ components, selected by the mixture-component sweep in \sifig{fig:si_gmm_kseep}, and the planning horizon is $T{=}20$ steps at 10\,Hz.
The \textit{adv} selector is implemented with a histogram-based gradient boosting classifier and a five-step temporal confirmation gate.
Additional \textit{adv}-selection features, diffusion losses, and reinforcement learning details are provided in the Supplementary Information.

Trajectory feasibility is measured from the \textit{adv} trajectory, with details provided in the Supplementary Section~\ref{si:kinematics}.
Let $a$ denote longitudinal acceleration and $j=\mathrm{d}a/\mathrm{d}t$ denote longitudinal jerk.
IP\% is computed per scenario as the fraction of time steps that exceed at least one physical feasibility bound: $|a|>7\,\mathrm{m/s^2}$~\cite{unece2021r157}, $|j|>12.65\,\mathrm{m/s^3}$~\cite{unece2021r157}, or $|a_{\mathrm{lat}}|>3.0\,\mathrm{m/s^2}$~\cite{unece2020grva7}.
The reported IP\% is averaged across scenarios, enabling cross-method comparison when rollout lengths differ.
Full percentile-level kinematic checks are reported in Supplementary Table~\ref{tab:si_kinematics}.


\section*{Data Availability}
The nuScenes dataset that we used to train CARS is publicly available at \url{https://www.nuscenes.org}. The RounD and AD4CHE dataset that we used for cross-dataset generalization experiment is publicly available at \url{https://levelxdata.com/round-dataset/} and can be requested at \url{https://lab.autozyx.com/ad4che.html}.

\section*{Code Availability}
The project website and the source code of the CARS framework is publicly available at \url{https://robosafe-lab.github.io/CARS}.

\section*{Acknowledgements}
This work was funded by UK Research and Innovation (UKRI) under the UK government's Horizon Europe funding guarantee [grant number EP/Z533464/1].

\section*{Author Contributions}
Y.X. developed the methodology, implemented the framework, conducted experiments, and wrote the manuscript.
H.Y. contributed to the visualization and experimental design.
Y.W. and Z.Z. oversaw the research project and guided the team.
Y.Z. provided data and cleaned the data.
X.Y and M.S.E. contributed to writing – review \& editing.
C.W. focused on conceptualization, supervision, funding, and editing.
All authors reviewed and approved the final version.

\section*{Competing Interests}
The authors declare no competing interests.

\bibliographystyle{naturemag}
\bibliography{reference}


\clearpage
\phantomsection
\addcontentsline{toc}{section}{Supplementary Information}

\setcounter{section}{0}
\setcounter{figure}{0}
\setcounter{table}{0}
\setcounter{equation}{0}
\renewcommand{\thesection}{S\arabic{section}}
\renewcommand{\theequation}{S\arabic{equation}}
\renewcommand{\thefigure}{S\arabic{figure}}
\renewcommand{\thetable}{S\arabic{table}}
\renewcommand{\theHsection}{S\arabic{section}}
\renewcommand{\theHequation}{S\arabic{equation}}
\renewcommand{\theHfigure}{S\arabic{figure}}
\renewcommand{\theHtable}{S\arabic{table}}
\setcounter{secnumdepth}{2}

\begin{center}
  {\Large\bfseries Supplementary Information}\\[6pt]
\end{center}
\vspace{12pt}

\setlength{\intextsep}{18pt plus 4pt minus 4pt}
\setlength{\textfloatsep}{18pt plus 4pt minus 4pt}
\setlength{\floatsep}{14pt plus 3pt minus 3pt}

\captionsetup[table]{font={small,sf}, labelfont=bf, labelsep=ncbar,
  singlelinecheck=off, justification=raggedright, skip=2pt}

\section{Lower-bound interpretation of FSM-attributed responsibility}
\label{si:proof_thm1}

This note formalizes the identifiability statement summarized in the main-text Methods (\nameref{sec:validity_verification}).
For a generated collision scenario $\xi$, let $\rho_{\mathrm{tgt}}$ denote the recorded geometric path of the target vehicle and let $\mathcal{T}_{\textit{adv}}$ denote the generated \textit{adv} trajectory.
For any target-control policy $\pi$, let $\mathcal{C}_{\pi}(\xi)$ be the replay outcome obtained by holding $\mathcal{T}_{\textit{adv}}$ fixed and applying $\pi$ to the target, and let $\mathrm{Collision}(\mathcal{C}_{\pi}(\xi))\in\{0,1\}$ indicate whether that replay still collides.
Let $\Pi$ be the set of admissible target-control policies satisfying the kinematic envelope, and let $\Pi_{\mathrm{lon}}\subseteq\Pi$ be the subset of policies that control only the target's longitudinal speed or braking along the fixed path $\rho_{\mathrm{tgt}}$.
Define
\begin{align}
\mathrm{Prev}(\xi)
&= \mathbf{1}\!\left[\exists\,\pi\in\Pi:
\mathrm{Collision}(\mathcal{C}_{\pi}(\xi))=0\right],\\
\mathrm{Prev}_{\mathrm{lon}}(\xi)
&= \mathbf{1}\!\left[\exists\,\pi\in\Pi_{\mathrm{lon}}:
\mathrm{Collision}(\mathcal{C}_{\pi}(\xi))=0\right],\\
\widehat{\mathrm{Prev}}_{\mathrm{FSM}}(\xi)
&= \mathbf{1}\!\left[
\mathrm{Collision}(\mathcal{C}_{\pi_{\mathrm{FSM}}}(\xi))=0\right],
\end{align}
where $\pi_{\mathrm{FSM}}$ is the FSM braking policy used in the path-constrained replay.
The claim is
\begin{equation}
\widehat{\mathrm{Prev}}_{\mathrm{FSM}}(\xi) \le \mathrm{Prev}_{\mathrm{lon}}(\xi) \le \mathrm{Prev}(\xi).
\label{eq:bound_chain}
\end{equation}
The argument proceeds in two parts: the chain of inequalities and the equality conditions.
We use three assumptions only when discussing equality, not for the lower-bound inequality itself:
\begin{itemize}
    \item \textbf{A1 (adversary-envelope validity).} The fixed \textit{adv} trajectory $\mathcal{T}_{\textit{adv}}$ respects the kinematic envelope assumed by the reference model.
    \item \textbf{A2 (path-constrained replay validity).} Evaluating longitudinal target control along $\rho_{\mathrm{tgt}}$ is an admissible approximation for the conflict class under study; i.e., the replay operator $\mathcal{C}_{\pi}$ represents a longitudinal response on the fixed encounter geometry.
    \item \textbf{A3 (FSM-completeness on the longitudinal envelope).} On the longitudinal conflict class addressed by UN\,R157, if any admissible longitudinal-only policy in $\Pi_{\mathrm{lon}}$ can avoid the collision, then the FSM policy also avoids it.
\end{itemize}

\paragraph{Proof of $\widehat{\mathrm{Prev}}_{\mathrm{FSM}}(\xi) \le \mathrm{Prev}_{\mathrm{lon}}(\xi)$.}
The FSM controller $\pi_{\mathrm{FSM}}$ is by construction a member of $\Pi_{\mathrm{lon}}$ (it brakes longitudinally and does not steer).
If $\widehat{\mathrm{Prev}}_{\mathrm{FSM}}(\xi)=1$, i.e.\ $\mathrm{Collision}(\mathcal{C}_{\pi_{\mathrm{FSM}}}(\xi))=0$, then $\pi_{\mathrm{FSM}} \in \Pi_{\mathrm{lon}}$ is a witness for the existential statement defining $\mathrm{Prev}_{\mathrm{lon}}(\xi)$, so $\mathrm{Prev}_{\mathrm{lon}}(\xi)=1$.
The contrapositive ($\mathrm{Prev}_{\mathrm{lon}}(\xi)=0 \Rightarrow \widehat{\mathrm{Prev}}_{\mathrm{FSM}}(\xi)=0$) holds because if no $\Pi_{\mathrm{lon}}$ controller can avoid the collision, then $\pi_{\mathrm{FSM}} \in \Pi_{\mathrm{lon}}$ in particular cannot.
Pointwise dominance of indicators yields the inequality.

\paragraph{Proof of $\mathrm{Prev}_{\mathrm{lon}}(\xi) \le \mathrm{Prev}(\xi)$.}
$\Pi_{\mathrm{lon}} \subseteq \Pi$ by definition.
Hence
\[
\{\exists\,\pi \in \Pi_{\mathrm{lon}} : \mathrm{Collision}(\mathcal{C}_{\pi}(\xi))=0\}
\subseteq
\{\exists\,\pi \in \Pi : \mathrm{Collision}(\mathcal{C}_{\pi}(\xi))=0\},
\]
so the indicator of the former is bounded above by that of the latter.

\paragraph{Equality of the first inequality under A1--A3.}
Assumption A3 (FSM-completeness) is precisely the implication $\mathrm{Prev}_{\mathrm{lon}}(\xi)=1 \Rightarrow \widehat{\mathrm{Prev}}_{\mathrm{FSM}}(\xi)=1$ on the longitudinal envelope addressed by UN\,R157.
Combined with $\widehat{\mathrm{Prev}}_{\mathrm{FSM}}(\xi) \le \mathrm{Prev}_{\mathrm{lon}}(\xi)$ above, this yields equality on that subset.
A1 ensures that FSM's evaluation is conducted on an \textit{adv} trajectory inside the kinematic envelope FSM itself assumes; A2 ensures that the path-constrained replay operator is the same one used to define $\Pi_{\mathrm{lon}}$.
Without A1 or A2, the regulatory interpretation and equality claim should not be invoked, although the lower-bound inequality remains valid whenever $\mathcal{C}_{\pi_{\mathrm{FSM}}}$ is still a member of the replay class represented by $\Pi_{\mathrm{lon}}$.

\paragraph{Equality of the second inequality.}
$\mathrm{Prev}_{\mathrm{lon}}(\xi)=\mathrm{Prev}(\xi)$ holds whenever $\mathrm{Prev}(\xi)=1 \Rightarrow \mathrm{Prev}_{\mathrm{lon}}(\xi)=1$, i.e.\ whenever the scenario's preventability is realizable through longitudinal control alone.
Scenarios that are preventable only through steering or combined longitudinal--lateral control fall in the strict-inequality regime.

\paragraph{Aggregate consequence.}
Summing pointwise inequalities over the corpus and dividing by $|\mathcal{S}_{\mathrm{col}}|$ preserves the inequalities. Hence
\begin{equation}
\hat r_{\mathrm{FSM}} \;=\; \frac{1}{|\mathcal{S}_{\mathrm{col}}|}\sum_{\xi}\widehat{\mathrm{Prev}}_{\mathrm{FSM}}(\xi) \;\le\; \frac{1}{|\mathcal{S}_{\mathrm{col}}|}\sum_{\xi}\mathrm{Prev}_{\mathrm{lon}}(\xi) \;\le\; \frac{1}{|\mathcal{S}_{\mathrm{col}}|}\sum_{\xi}\mathrm{Prev}(\xi).
\end{equation}
The reported $\hat r_{\mathrm{FSM}}=88.7\%$ is therefore a deterministic lower bound on the true preventability fraction across the corpus, regardless of which subset of scenarios is $\Pi_{\mathrm{lon}}$-realizable, and is exact on the longitudinal-realizability subset under A1--A3.

\paragraph{Bias under relaxation.}
The argument relies on no probabilistic structure (no scenario sampling distribution, no expectation over noise); the bound is deterministic and pointwise.
The lower-bound conclusion in~\eqref{eq:bound_chain} requires only that $\mathcal{C}_{\pi_{\mathrm{FSM}}}$ corresponds to a longitudinal controller in $\Pi_{\mathrm{lon}}$ replayed on $\rho_{\mathrm{tgt}}$; it remains valid under wider conditions than the equality.
If A1 or A2 fails, the regulatory interpretation and equality claim should not be invoked; the lower-bound claim still holds as long as the FSM replay remains a longitudinal controller in $\Pi_{\mathrm{lon}}$.
Relaxing A3 alone changes only the equality on the $\Pi_{\mathrm{lon}}$-realizable subset, leaving the lower-bound conclusion intact.

\section{Adversarial Agent Selection: Feature Vector}
\label{si:adv_features}

The 13-dimensional feature vector $\boldsymbol{\phi}_{i,t}$ for each surrounding agent $i$ relative to \textit{tgt} is defined at control step $t$ as follows.
Let \textit{tgt} have position $\mathbf{p}_{\textit{tgt},t}$, velocity $\mathbf{v}_{\textit{tgt},t}$, speed $v_{\textit{tgt},t}$, heading $\theta_{\textit{tgt},t}$, and length $L_{\textit{tgt}}$.
Let $\mathbf{e}^{\parallel}_{\textit{tgt},t}=(\cos\theta_{\textit{tgt},t},\sin\theta_{\textit{tgt},t})^\top$ and $\mathbf{e}^{\perp}_{\textit{tgt},t}=(-\sin\theta_{\textit{tgt},t},\cos\theta_{\textit{tgt},t})^\top$ denote its longitudinal and lateral unit vectors.
For each surrounding agent $i$ with state $(\mathbf{p}_{i,t},\mathbf{v}_{i,t},v_{i,t},\theta_{i,t},L_i)$, define $\mathbf{e}^{\parallel}_{i,t}=(\cos\theta_{i,t},\sin\theta_{i,t})^\top$ and $\Delta\mathbf{p}_{i,t}=\mathbf{p}_{i,t}-\mathbf{p}_{\textit{tgt},t}$.

\begin{equation}
\begin{array}{ll@{\qquad}ll}
d_{i,t} = \|\Delta\mathbf{p}_{i,t}\|,              &
& \Delta v_{i,t} = v_{i,t} - v_{\textit{tgt},t},                           \\[3pt]
c_{i,t} = \cos(\theta_{i,t} - \theta_{\textit{tgt},t}),            &
& s_{i,t} = \Delta\mathbf{p}_{i,t}^{\top}\mathbf{e}^{\parallel}_{\textit{tgt},t},       \\[3pt]
l_{i,t} = \Delta\mathbf{p}_{i,t}^{\top}\mathbf{e}^{\perp}_{\textit{tgt},t}, &
& \dot{s}_{i,t} = (\mathbf{v}_{\textit{tgt},t} - \mathbf{v}_{i,t})^{\top}\mathbf{e}^{\parallel}_{\textit{tgt},t}, \\[3pt]
\dot{l}_{i,t} = (\mathbf{v}_{\textit{tgt},t} - \mathbf{v}_{i,t})^{\top}\mathbf{e}^{\perp}_{\textit{tgt},t}, &
& g_{i,t} = s_{i,t} - \tfrac{L_{\textit{tgt}} + L_i}{2} - b,       \\[3pt]
\mathrm{TTC}_{i,t} = g_{i,t} / \max(\dot{s}_{i,t},\, \epsilon), &
& \beta_{i,t} =
\dfrac{(\mathbf{p}_{\textit{tgt},t}-\mathbf{p}_{i,t})^{\top}\mathbf{e}^{\parallel}_{i,t}}
{\max(\|\mathbf{p}_{\textit{tgt},t}-\mathbf{p}_{i,t}\|,\epsilon)}, \\[8pt]
\dot{d}_{i,t}^{(5)} = (d_{i,t} - d_{i,t-5}) / (5\Delta t), &
& \Delta|l|_{i,t}^{(5)} = |l_{i,t}| - |l_{i,t-5}|,
\end{array}
\label{eq:si_adv_features}
\end{equation}
where $d_{i,t}$ is Euclidean distance, $\Delta v_{i,t}$ relative speed, $c_{i,t}$ heading alignment, $s_{i,t}$/$l_{i,t}$ longitudinal/lateral offset, $\dot{s}_{i,t}$/$\dot{l}_{i,t}$ longitudinal/lateral closing speed, $g_{i,t}$ bumper-to-bumper gap ($b$ is a bumper margin), $\mathrm{TTC}_{i,t}$ time to collision, $\beta_{i,t}$ the cosine of the bearing angle from agent $i$ toward \textit{tgt}, $\dot{d}_{i,t}^{(5)}$ the distance change rate over the most recent 5 steps, and $\Delta|l|_{i,t}^{(5)}$ the change in absolute lateral offset. Together with the agent speed $v_{i,t}$, this yields a 13-dimensional feature vector $\boldsymbol{\phi}_{i,t}$.

\section{Diffusion Model: Forward Process}
\label{si:forward_diffusion}

The forward diffusion process defines a Markov chain that gradually corrupts the clean action sequence $\tau_a^0$ into Gaussian noise over $R$ diffusion steps. Each forward transition $q(\tau_a^{r}\mid\tau_a^{r-1})$ is a fixed Gaussian kernel:
\begin{equation}
q(\tau_a^r \mid \tau_a^{r-1}) =
\mathcal{N}\!\big(\sqrt{\alpha_r}\,\tau_a^{r-1},\, \beta_r \mathbf{I}\big),
\label{eq:si_forward}
\end{equation}
where $\{\beta_r\}_{r=1}^R$ is a predefined variance schedule, $\alpha_r = 1 - \beta_r$, and
$\bar{\alpha}_r = \prod_{u=1}^r \alpha_u$.
The resulting marginal distribution admits the closed-form:
\begin{equation}
\tau_a^r = \sqrt{\bar{\alpha}_r}\,\tau_a^0
+ \sqrt{1 - \bar{\alpha}_r}\,\boldsymbol{\epsilon},
\quad \boldsymbol{\epsilon} \sim \mathcal{N}(\mathbf{0}, \mathbf{I}).
\label{eq:si_marginal}
\end{equation}

\section{GMDM Loss Components}
\label{si:loss_components}

The total loss in \eqref{eq:total_loss} comprises three terms:
$\mathcal{L}_{\mathrm{total}} = \mathcal{L}_{\mathrm{NLL}} - \lambda_H \mathcal{H}_\pi + \lambda_{\mathrm{repel}} \mathcal{L}_{\mathrm{repel}}$.

\paragraph{(a) Negative log-likelihood.}
The negative log-likelihood term is
\begin{equation}
\mathcal{L}_{\mathrm{NLL}}(\theta)
=
-\sum_{h=1}^{T}
\log p_{\theta}\!\left(\tau_{a,h}^0 \mid \tau_a^r, \mathbf{c}, r\right).
\label{eq:si_nll}
\end{equation}

\paragraph{(b) Entropy regularization.}
The mixture-weight entropy term is
\begin{equation}
\mathcal{H}_{\pi}
=
\sum_{h=1}^{T}
\left(
-\sum_{k=1}^{K} \pi_{h,k}\log(\pi_{h,k})
\right),
\label{eq:si_entropy}
\end{equation}
where $K$ denotes the number of Gaussian components and $\pi_{h,k}$ is the weight of component $k$ at horizon step $h$.

\paragraph{(c) Component repulsion.}
The component-repulsion term is
\begin{equation}
\mathcal{L}_{\mathrm{repel}}
=
\frac{1}{T}
\sum_{h=1}^{T}
\frac{1}{K(K-1)}
\sum_{k\neq \ell}
\exp\!\left(
-\frac{\|\mu_{h,k}-\mu_{h,\ell}\|_2^2}{s^2}
\right),
\label{eq:si_repel}
\end{equation}
where $s$ controls the distance scale for component separation.

\section{PPO Fine-Tuning: Detailed Formulation}
\label{si:ppo}

The reverse denoising process is formulated as a multi-step Markov decision process.
At diffusion step $r$, the state and policy are defined as
\begin{equation}
s_r = (\tau_a^r,\, r,\, \mathbf{c}),
\quad
\pi_\theta(\tau_a^{r-1} \mid s_r)
=
p_\theta(\tau_a^{r-1} \mid \tau_a^r, \mathbf{c}).
\label{eq:si_mdp}
\end{equation}

The objective is to maximize
\begin{equation}
J(\theta)
=
\mathbb{E}_{\pi_\theta}\!\big[
\mathcal{R}(\tau_a^0, \mathbf{c})
\big].
\label{eq:si_ppo_obj}
\end{equation}

The policy gradient across $R$ denoising steps is
\begin{equation}
\nabla_\theta J
=
\mathbb{E}
\left[
\sum_{r=1}^{R}
\nabla_\theta
\log p_\theta(\tau_a^{r-1} \mid \tau_a^r, \mathbf{c})
\;
\mathcal{R}(\tau_a^0,\mathbf{c})
\right].
\label{eq:si_policy_grad}
\end{equation}

The clipped PPO surrogate loss is
\begin{equation}
\mathcal{L}_{\pi}(\theta)
= -\mathbb{E}\!\left[\sum_{r=1}^{R}
\min\!\Big(\rho_\theta^{(r)}\hat{A}^{(r)},\;
\mathrm{clip}\!\big(\rho_\theta^{(r)},\,1{-}\epsilon,\,1{+}\epsilon\big)
\hat{A}^{(r)}\Big)\right],
\label{eq:si_ppo_loss}
\end{equation}
where $\rho_\theta^{(r)} = p_\theta(\tau_a^{r-1} \mid \tau_a^r, \mathbf{c}) \,/\, p_{\theta_{\mathrm{old}}}(\tau_a^{r-1} \mid \tau_a^r, \mathbf{c})$ is the importance ratio, $\hat{A}^{(r)}$ is the generalized advantage estimate at denoising step $r$, and $\epsilon$ is the clipping coefficient.

\paragraph{Reward components.}
Let $d^{\mathrm{lon}}_0$ and $d^{\mathrm{lon}}_1$ denote the longitudinal component of the \textit{adv}--\textit{tgt} relative position projected onto the \textit{tgt} heading at the start and end of an $n$-step shaping window, and $\ell_0$, $\ell_1$ the corresponding lateral components.
The individual reward terms in $R_{\mathrm{total}}$ (\eqref{eq:reward_total}) are:
\begin{align}
& R_{\mathrm{prog}} = \mathrm{clip}\!\left(
    \frac{d^{\mathrm{lon}}_0 - d^{\mathrm{lon}}_1}{d^{\mathrm{lon}}_0 + \nu},\;
    -0.5,\; 1\right),                                                               \nonumber\\[4pt]
& R_{\ell} = \begin{cases}
    \phantom{-0.5\,}\exp(-\ell_1/\sigma_\ell) & \text{if } \ell_0 > \ell_1, \\
    -0.5\,\exp(-\ell_1/\sigma_\ell)            & \text{otherwise,}
    \end{cases}                                                                     \nonumber\\[4pt]
& R_{\ell c} = \mathrm{clip}\!\left(
    \frac{\ell_0 - \ell_1}{\Delta t_{\mathrm{eff}}\,(\ell_0 + \nu)},\;
    -0.5,\; 1\right),
\label{eq:si_reward_components}
\end{align}
where $\sigma_\ell$ is a lateral decay scale, $\Delta t_{\mathrm{eff}}$ is the shaping-window duration, and $\nu > 0$ is a small constant for numerical stability.
The clip at $-0.5$ allows the reward to penalize diverging motion up to a bounded magnitude, so the \textit{adv} avoids pathological retreat without being driven off-manifold by a sharp negative gradient.

\section{FSM: Full Specification}
\label{si:fsm_full_spec}
\label{si:fsm_distances}

The implementation follows UN Regulation No.~157~\cite{unece2021r157}.
The regulatory parameters are the reaction time $\tau{=}0.75\,$s, the \textit{tgt} comfort deceleration $b_{\textit{tgt},\mathrm{comf}}{=}4\,$m/s$^2$, the \textit{tgt} maximum deceleration $b_{\textit{tgt},\max}{=}6\,$m/s$^2$, the assumed \textit{adv} maximum deceleration $b_{\textit{adv},\max}{=}7\,$m/s$^2$, the minimum standstill distance $d_{\min}{=}2\,$m, and the jerk limit $J{=}12.65\,$m/s$^3$.
Here $b_{\textit{tgt}}$ denotes the commanded positive deceleration magnitude before the reaction-delay and jerk-ramp implementation.
$v_{\textit{adv}}$ and $v_{\textit{tgt}}$ denote longitudinal speeds; $a_{\textit{tgt}}$ is the signed longitudinal acceleration of \textit{tgt}.
$d_{\mathrm{lon}}$ and $d_{\mathrm{lat}}$ denote the longitudinal and lateral separations in the \textit{tgt}-aligned frame, $v_{\textit{adv},\mathrm{lat}}$ is the \textit{adv} lateral approach speed toward \textit{tgt}, and $L_{\textit{adv}}$ and $L_{\textit{tgt}}$ denote vehicle lengths.

\paragraph{Lateral safety pre-check.}
Lateral risk is identified only when all four conditions hold; failure of any condition causes FSM to skip the longitudinal fuzzy stage and apply no braking:
\begin{equation}
\begin{array}{l}
\text{(a)}\;\; d_{\mathrm{lon}} > \tfrac{1}{2}(L_{\textit{tgt}} + L_{\textit{adv}})\quad\text{(\textit{adv} rear ahead of \textit{tgt} front)},\\[2pt]
\text{(b)}\;\; v_{\textit{adv},\mathrm{lat}} > 0\quad\text{(\textit{adv} moving toward \textit{tgt} laterally)},\\[2pt]
\text{(c)}\;\; v_{\textit{tgt}} > v_{\textit{adv}}\quad\text{(\textit{tgt} closing on \textit{adv} longitudinally)},\\[2pt]
\text{(d)}\;\; \dfrac{d_{\mathrm{lat}}}{v_{\textit{adv},\mathrm{lat}}} <
              \dfrac{d_{\mathrm{lon}} + L_{\textit{tgt}} + L_{\textit{adv}}}{v_{\textit{tgt}} - v_{\textit{adv}}} + 0.1.
\end{array}
\label{eq:si_lateral_check}
\end{equation}

\paragraph{Fuzzy membership.}
PFS and CFS share a saturated linear membership $\mu \in [0,1]$:
\begin{equation}
\mu(x;\, x_{\mathrm{safe}},\, x_{\mathrm{unsafe}}) = \mathrm{clamp}\!\left(\frac{x_{\mathrm{safe}} - x}{x_{\mathrm{safe}} - x_{\mathrm{unsafe}}},\; 0,\; 1\right).
\label{eq:si_fuzzy_mu}
\end{equation}

\paragraph{PFS.}
\begin{equation}
\begin{aligned}
d_{\mathrm{safe}}^{\mathrm{PFS}} &= v_{\textit{tgt}}\tau + \frac{v_{\textit{tgt}}^2}{2b_{\textit{tgt},\mathrm{comf}}} - \frac{v_{\textit{adv}}^2}{2b_{\textit{adv},\max}} + d_{\min},\\[2pt]
d_{\mathrm{unsafe}}^{\mathrm{PFS}} &= v_{\textit{tgt}}\tau + \frac{v_{\textit{tgt}}^2}{2b_{\textit{tgt},\max}}  - \frac{v_{\textit{adv}}^2}{2b_{\textit{adv},\max}},\\[2pt]
\mathrm{PFS} &= \mu(d_{\mathrm{lon}} - d_{\min};\, d_{\mathrm{safe}}^{\mathrm{PFS}},\, d_{\mathrm{unsafe}}^{\mathrm{PFS}}).
\end{aligned}
\label{eq:si_pfs}
\end{equation}
The $d_{\min}$ safety distance appears both as a constant offset in $d_{\mathrm{safe}}^{\mathrm{PFS}}$ and as a shift on the membership input.

\paragraph{CFS.}
CFS is computed only when $v_{\textit{tgt}} > v_{\textit{adv}}$; otherwise $\mathrm{CFS}{=}0$. Define $a_{\textit{tgt}}' = \max(a_{\textit{tgt}},\, -b_{\textit{tgt},\mathrm{comf}})$ and $v_{\textit{tgt}}^* = v_{\textit{tgt}} + a_{\textit{tgt}}'\,\tau$. Two branches:
\begin{equation}
\bigl(d_{\mathrm{safe}}^{\mathrm{CFS}},\, d_{\mathrm{unsafe}}^{\mathrm{CFS}}\bigr) =
\begin{cases}
\Bigl(\dfrac{(v_{\textit{tgt}} - v_{\textit{adv}})^2}{2|a_{\textit{tgt}}'|},\,\dfrac{(v_{\textit{tgt}} - v_{\textit{adv}})^2}{2|a_{\textit{tgt}}'|}\Bigr),
  & v_{\textit{tgt}}^* \le v_{\textit{adv}}, \\[8pt]
\Bigl(d_{\mathrm{new}} + \dfrac{(v_{\textit{tgt}}^* - v_{\textit{adv}})^2}{2b_{\textit{tgt},\mathrm{comf}}},\,
      d_{\mathrm{new}} + \dfrac{(v_{\textit{tgt}}^* - v_{\textit{adv}})^2}{2b_{\textit{tgt},\max}}\Bigr),
  & v_{\textit{tgt}}^* > v_{\textit{adv}},
\end{cases}
\label{eq:si_cfs}
\end{equation}
where $d_{\mathrm{new}} = \bigl((v_{\textit{tgt}} + v_{\textit{tgt}}^*)/2 - v_{\textit{adv}}\bigr)\tau$. Then $\mathrm{CFS} = \mu(d_{\mathrm{lon}};\, d_{\mathrm{safe}}^{\mathrm{CFS}},\, d_{\mathrm{unsafe}}^{\mathrm{CFS}})$. The first branch handles the case where the \textit{tgt} would already be slower than the \textit{adv} after the reaction window, in which case $d_{\mathrm{safe}} = d_{\mathrm{unsafe}}$ degenerates the membership to a step function under UN R157.

\paragraph{Braking response.}
\begin{equation}
b_{\textit{tgt}} =
\begin{cases}
\mathrm{CFS}\,(b_{\textit{tgt},\max} - b_{\textit{tgt},\mathrm{comf}}) + b_{\textit{tgt},\mathrm{comf}}, & \mathrm{CFS} > 0, \\[2pt]
\mathrm{PFS}\, b_{\textit{tgt},\mathrm{comf}}, & \mathrm{CFS} = 0.
\end{cases}
\label{eq:si_fsm_reaction}
\end{equation}
When CFS is non-zero, the reaction is at least $b_{\textit{tgt},\mathrm{comf}}$ and rises linearly to $b_{\textit{tgt},\max}$ as CFS approaches one. When CFS vanishes, PFS scales the comfort deceleration only.

\paragraph{Reaction-time and jerk-limited ramp.}
The deceleration is implemented after a delay $\tau$, then increases at the constant jerk rate $J=12.65$\,m/s$^3$ until it reaches $b_{\textit{tgt}}$:
\begin{equation}
b_{\mathrm{actual}}(t) =
\begin{cases}
0, & t < t_{\mathrm{trig}} + \tau, \\[2pt]
\min\!\bigl(b_{\mathrm{actual}}(t{-}\Delta t) + J\Delta t,\; b_{\textit{tgt}}(t)\bigr), & t \ge t_{\mathrm{trig}} + \tau,
\end{cases}
\label{eq:si_fsm_ramp}
\end{equation}
where $t_{\mathrm{trig}}$ is the first step at which $b_{\textit{tgt}} > 0$. The path-constrained replay in \cref{si:roller_coaster} integrates $b_{\mathrm{actual}}$ as the FSM-commanded deceleration $a_{\mathrm{FSM}}(t)$.

\section{Path-Constrained Re-Simulation}
\label{si:roller_coaster}

The \textit{tgt} follows its original recorded path, parameterized by arc length $s$; only its speed is modified by the reference models.
Let $v_{\mathrm{orig}}(t)$ denote the \textit{tgt}'s original speed at step $t$.
Let $a_{\mathrm{ref}}(t)\geq 0$ denote the reference-model deceleration command along this fixed path; for FSM, $a_{\mathrm{ref}}(t)=a_{\mathrm{FSM}}(t)$ from \cref{eq:si_fsm_ramp}, which already includes the reaction delay and jerk-limited ramp.
A cumulative speed deficit $\Delta v$ then accumulates this commanded deceleration:
\begin{equation}
\left\{
\begin{array}{l}
\Delta v(0)=0,\\[3pt]
\Delta v(t) = \Delta v(t{-}1) + a_{\mathrm{ref}}(t)\cdot\Delta t, \\[3pt]
v(t) = \max\!\bigl(0,\; v_{\mathrm{orig}}(t) - \Delta v(t)\bigr), \\[3pt]
s(t{+}1) = s(t) + v(t)\cdot\Delta t,
\end{array}
\right.
\label{eq:si_roller_coaster}
\end{equation}
where $\Delta t$ is the simulation time step.
When $\Delta v = 0$, the re-simulated trajectory reproduces the original recording exactly.

\section{Gaussian Mixture Component Selection}
\label{si:gmm_kseep}

\begin{figure}[H]
  \centering
  \includegraphics[width=0.55\textwidth]{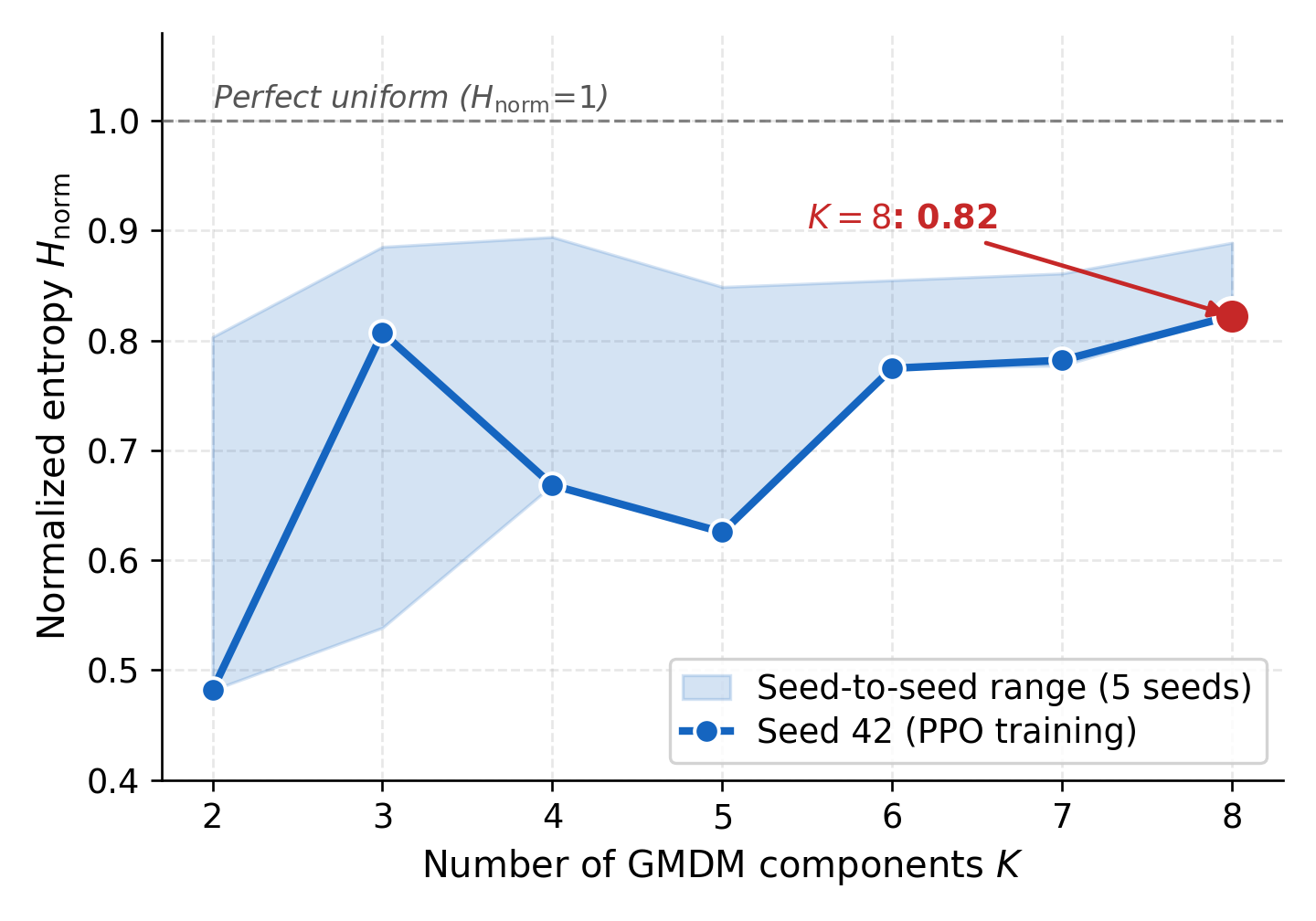}
  \caption{Normalized entropy $H_\mathrm{norm}$ of the GMDM mixture weights at convergence, as a function of the number of Gaussian components $K$. The dark line is the seed used for PPO training (seed 42) and the light envelope is the range across five independent seeds. $K{=}8$ achieves the highest single-seed utilization ($H_\mathrm{norm}{=}0.82$, highlighted in red) and is used throughout the paper.}
  \label{fig:si_gmm_kseep}
\end{figure}

We select the number of Gaussian mixture components in the GMDM by training models with $K \in \{2,3,4,5,6,7,8\}$ and measuring the normalized mixture-weight entropy $H_\mathrm{norm}$ at convergence. \Cref{fig:si_gmm_kseep} reports the sweep across five independent seeds. We use $K{=}8$ in the paper because it achieves the highest single-seed utilization on the PPO-training seed ($H_\mathrm{norm}{=}0.82$) and the narrowest seed-to-seed spread among the tested values, indicating reliable mixture-component usage.

\section{Statistical Robustness across Random Seeds}
\label{si:seed_robustness}

\begin{figure}[H]
  \centering
  \begin{subfigure}[t]{0.24\textwidth}
    \paneltitle{a}{Collision Episodes}
    \includegraphics[width=\linewidth]{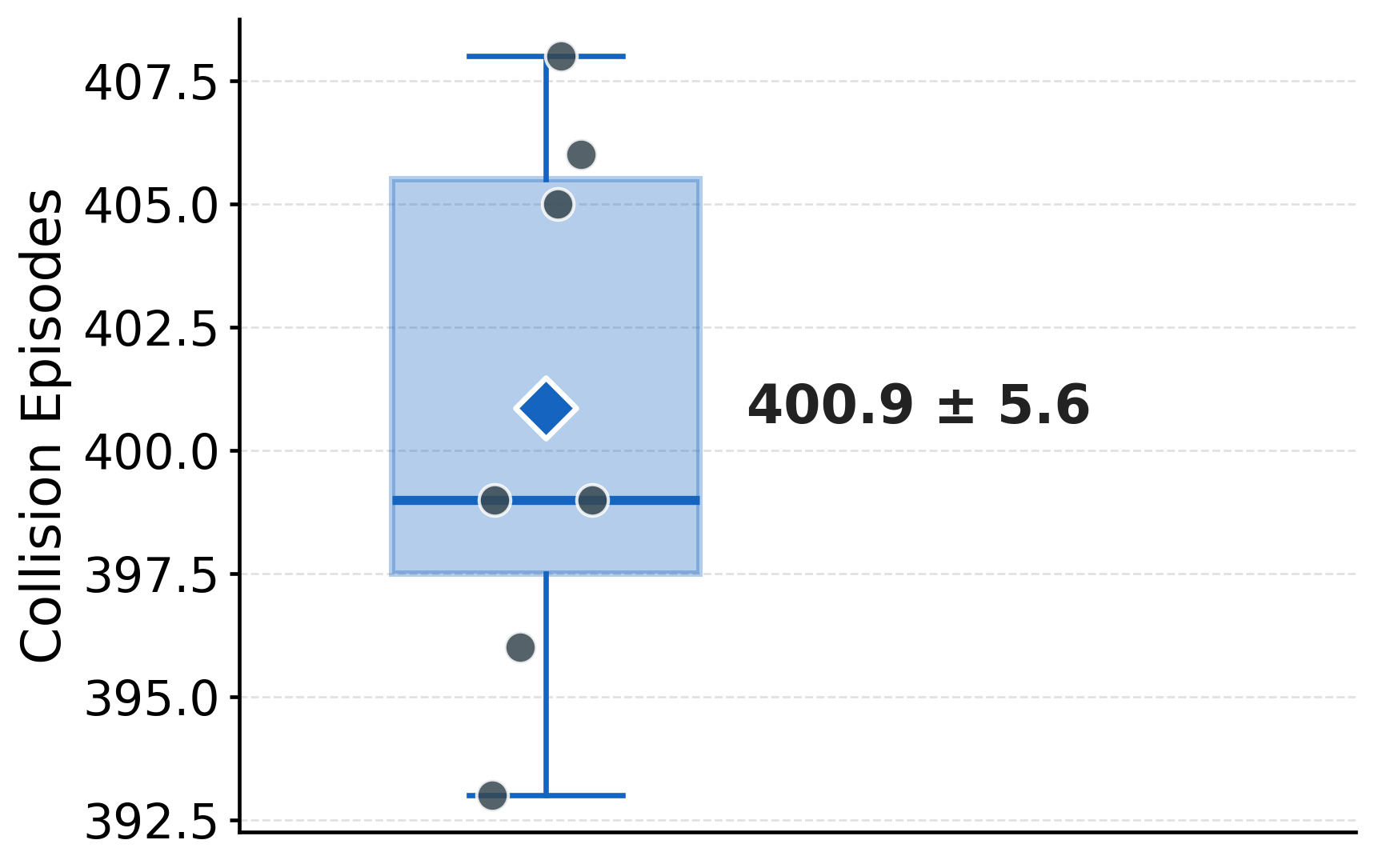}
  \end{subfigure}\hfill
  \begin{subfigure}[t]{0.24\textwidth}
    \paneltitle{b}{FSM attribution (\%)}
    \includegraphics[width=\linewidth]{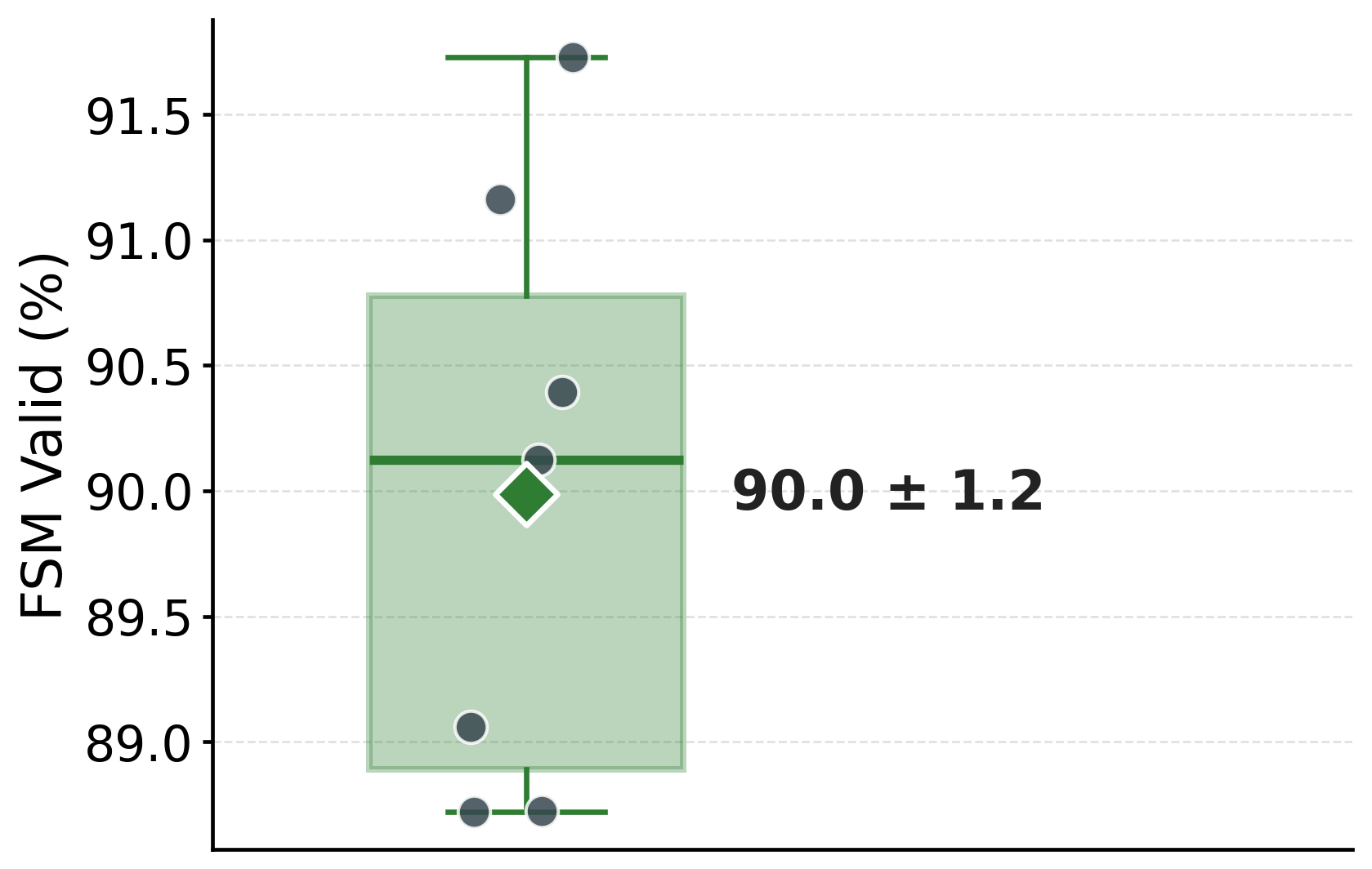}
  \end{subfigure}\hfill
  \begin{subfigure}[t]{0.24\textwidth}
    \paneltitle{c}{$H_\mathrm{crit}$}
    \includegraphics[width=\linewidth]{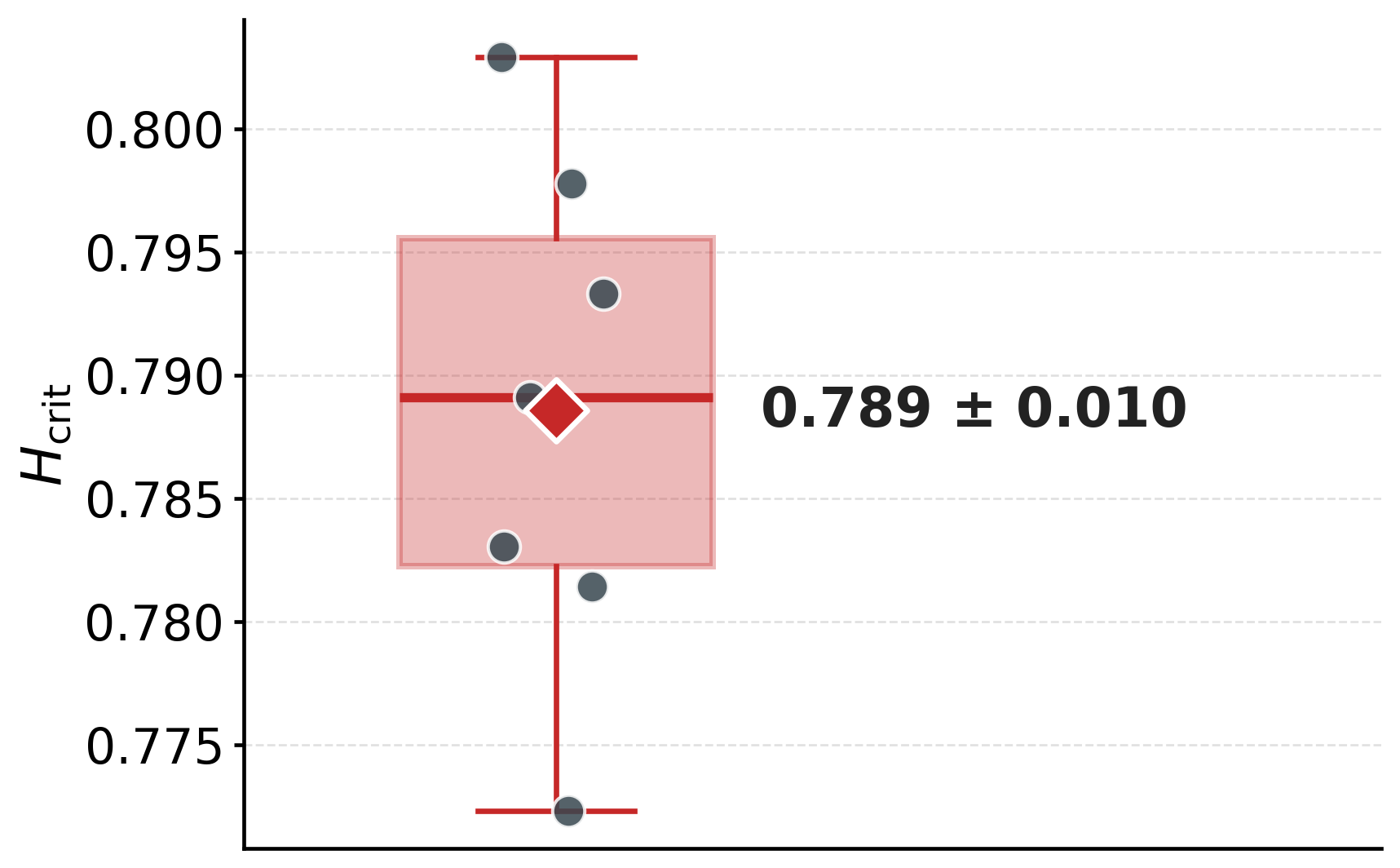}
  \end{subfigure}\hfill
  \begin{subfigure}[t]{0.24\textwidth}
    \paneltitle{d}{TTC$_\mathrm{min}$ median (s)}
    \includegraphics[width=\linewidth]{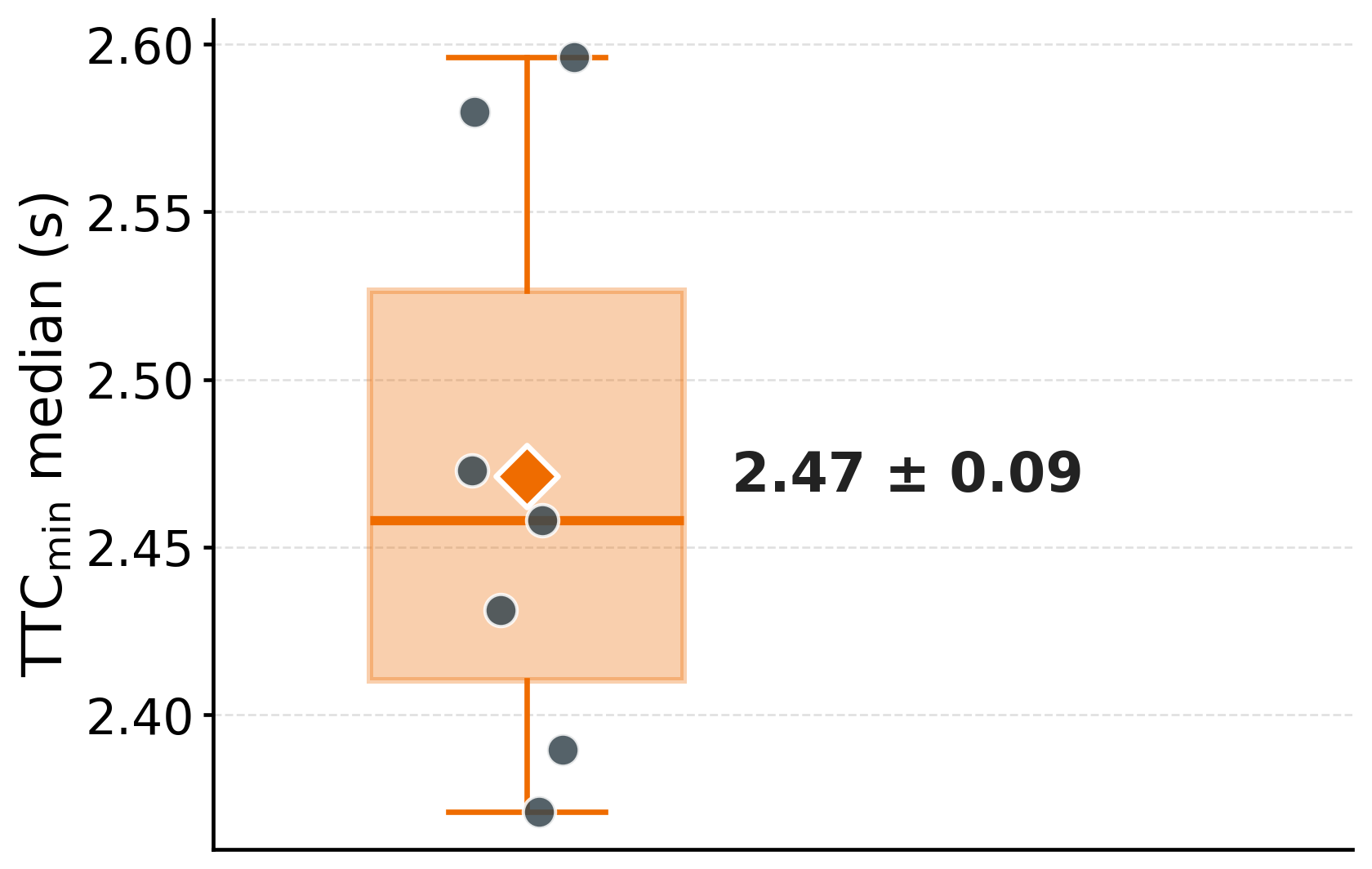}
  \end{subfigure}
  \caption{Statistical robustness across seven random seeds. Each panel shows the distribution of a key metric: box plot (median and interquartile range), individual seed values (dots), and mean (diamond).}
  \label{fig:si_seed_robustness}
\end{figure}

We repeat the closed-loop rollout with six additional seeds (seven seeds in total) using the same PPO-fine-tuned \textit{adv} policy and the same nuScenes scene--agent pairs. \Cref{fig:si_seed_robustness} reports collision count, FSM attribution, $H_{\mathrm{crit}}$, and median TTC$_\mathrm{min}$ across seeds. $H_{\mathrm{crit}}$ is computed on each seed's FSM-preventable subset, matching the definition used in main-text Table~\ref{tab:main_results}.

\section{CARS Scenario Quality Stratified by FSM Criticality}
\label{si:main_results}

Supplementary Tables~\ref{si:tab:main_results}--\ref{si:tab:feasible_criticality} provide the tier-level statistics underlying the severity-diversity and feasibility analyses in the main text. Supplementary Table~\ref{si:tab:main_results} reports per-tier scenario quality for FSM-preventable CARS collisions. Supplementary Table~\ref{si:tab:criticality_distribution} reports the Hard/Medium/Easy distribution used to compute $H_\mathrm{crit}$ across methods and transfer settings. Supplementary Table~\ref{si:tab:feasible_criticality} recomputes $H_\mathrm{crit}$ after applying stricter kinematic-feasibility requirements.

\section{Adversary Classifier Evaluation}
\label{si:dt_evaluation}

Supplementary Table~\ref{si:tab:dt_evaluation} reports binary-classification and scene-level ranking metrics for the context-aware \textit{adv} selector on the validation split.

\section{Cross-Reference Comparison}
\label{si:validity}

Supplementary Table~\ref{si:tab:validity} compares FSM, CC-JP, and RSS on the same 408 CARS collision scenarios on nuScenes. Quantities restricted to unpreventable cases are computed separately for each reference model's unpreventable subset.

\section{Full \textit{adv} Trajectory Kinematics}
\label{si:kinematics}

Supplementary Table~\ref{tab:si_kinematics} reports the \textit{adv} trajectory kinematics underlying the IP\% column in main-text Table~\ref{tab:main_results}.

Higher-order kinematic estimation from discretely sampled position is intrinsically noise-amplifying: triple finite differencing of cm-precision position at $\Delta t = 0.1$\,s amplifies measurement noise by $\sim 10^3$, producing a physically meaningless jerk floor on the order of $20\,\mathrm{m/s^3}$ even for smooth motion. This is an estimator-level requirement and is independent of any subsequent feasibility evaluation. We therefore apply a Savitzky-Golay filter (window 7 frames at 10\,Hz, cubic polynomial) to the position trace before differencing for all baselines; the table reports the mean of per-scenario maxima and 95th-percentile values.

The RounD entry additionally subtracts the geometric centripetal floor $v^2/r_\mathrm{local}$ from $|a_\mathrm{lat}|$ so that normal cornering on the fixed roundabout geometry is not counted as a lateral violation.

We write $j=\mathrm{d}a/\mathrm{d}t$ for longitudinal jerk.
IP\% uses the bounds $|a|\le 7\,\mathrm{m/s^2}$~\cite{unece2021r157}, $|j|\le 12.65\,\mathrm{m/s^3}$~\cite{unece2021r157}, and $|a_\mathrm{lat}|\le 3.0\,\mathrm{m/s^2}$~\cite{unece2020grva7}.

\clearpage
\section*{Supplementary Tables}

\begin{table}[H]
\centering
\caption{CARS scenario quality by FSM criticality tier on nuScenes.}
\label{si:tab:main_results}
\renewcommand{\arraystretch}{1.25}
\footnotesize\sffamily
\setlength{\tabcolsep}{4pt}
\begin{tabular}{lcccc}
\toprule
\textbf{Metric} & \textbf{Hard} & \textbf{Medium} & \textbf{Easy} & \textbf{Overall} \\
\midrule
Episodes (n, \%)                          & 26 (7.2) & 207 (57.2) & 129 (35.6) & 362 \\
TTC$_\mathrm{min}$ median (s) $\downarrow$ & 0.61  & 2.39    & 3.25   & 2.72 \\
BD$^+$\% $\uparrow$                        & 19.2  & 15.9 & 10.1 & 14.1 \\
$\bar{d}_\mathrm{min}$ (m) $\downarrow$    & 1.49  & 3.49    & 4.64   & 3.76 \\
\bottomrule
\end{tabular}
\end{table}

\begin{table}[H]
\centering
\caption{FSM criticality tier distribution across methods.}
\label{si:tab:criticality_distribution}
\renewcommand{\arraystretch}{1.25}
\footnotesize\sffamily
\setlength{\tabcolsep}{5pt}
\begin{tabular}{lccccc}
\toprule
\textbf{Method / Configuration} & $\boldsymbol{N}$ & \textbf{Hard\%} & \textbf{Medium\%} & \textbf{Easy\%} & $\boldsymbol{H_\mathrm{crit}}$ \\
\midrule
\multicolumn{6}{l}{\textit{Baselines (nuScenes)}} \\
STRIVE~\cite{rempe2022generating}      &  30 & 73.3 & 26.7 &  0.0 & 0.528 \\
SafeSim~\cite{chang2024safe}           &  13 & 46.2 & 53.8 &  0.0 & 0.628 \\
Bezier-CAT~\cite{zhang2023cat}         & 121 & 91.7 &  0.0 &  8.3 & 0.260 \\
\midrule
\multicolumn{6}{l}{\textit{Ours (nuScenes)}} \\
CARS                                    & 362 &  7.2 & 57.2 & 35.6 & \textbf{0.798} \\
CARS ($K{=}1$ \textit{adv})             &  28 & 35.7 & 57.1 &  7.1 & 0.797 \\
\midrule
\multicolumn{6}{l}{\textit{ADS-planner robustness (fixed CARS \textit{adv})}} \\
One-component diffusion planner                  & 158 &  8.9 & 53.8 & 37.3 & 0.834 \\
CTG planner~\cite{zhong2023guided}               & 271 &  6.6 & 64.9 & 28.4 & 0.745 \\
\midrule
\multicolumn{6}{l}{\textit{Cross-dataset generalization}} \\
CARS on AD4CHE~\cite{zhang2023ad4che}   & 359 &  8.6 & 66.9 & 24.5 & 0.751 \\
CARS on RounD~\cite{krajewski2020round} & 533 & 15.8 & 69.0 & 15.2 & 0.759 \\
\bottomrule
\end{tabular}
\end{table}

\begin{table}[H]
\centering
\caption{FSM-preventable severity diversity after applying kinematic feasibility requirements. 
$N_{0}$ and $H_{0}$ additionally require zero per-step kinematic violation ($\mathrm{IP}=0$). $N_{0.10}$ and $H_{0.10}$ allow at most 10\% violating time steps. 
}
\label{si:tab:feasible_criticality}
\renewcommand{\arraystretch}{1.25}
\footnotesize\sffamily
\setlength{\tabcolsep}{5pt}
\begin{tabular}{lcccccc}
\toprule
\textbf{Method / Configuration} & $\boldsymbol{N_\mathrm{prev}}$ & $\boldsymbol{H_\mathrm{crit}}$ & $\boldsymbol{N_0}$ & $\boldsymbol{H_0}$ & $\boldsymbol{N_{0.10}}$ & $\boldsymbol{H_{0.10}}$ \\
\midrule
CARS                                    & 362 & 0.798 & 352 & 0.796 & 362 & 0.798 \\
CARS ($K{=}1$ \textit{adv})     &  28 & 0.797 &   7 & 0.373 &   9 & 0.622 \\
\midrule
STRIVE~\cite{rempe2022generating}       &  30 & 0.528 &   3 & 0.000 &   4 & 0.000 \\
SafeSim~\cite{chang2024safe}            &  13 & 0.628 &   0 & NA    &   8 & 0.512 \\
Bezier-CAT~\cite{zhang2023cat}          & 121 & 0.260 &   0 & NA    &   0 & NA \\
\bottomrule
\end{tabular}
\end{table}

\begin{table}[H]
\centering
\caption{Adversary classifier evaluation on the validation split (62 scenes, 2{,}131 agent-level samples, 135 \textit{tgt}--\textit{adv} pairs).}
\label{si:tab:dt_evaluation}
\renewcommand{\arraystretch}{1.25}
\footnotesize\sffamily
\setlength{\tabcolsep}{8pt}
\begin{tabular}{lc}
\toprule
\textbf{Metric} & \textbf{Value} \\
\midrule
\multicolumn{2}{l}{\textit{Binary classification (per class)}} \\
Precision / Recall / F1 (adversary)     & 0.75 / 0.79 / 0.77 \\
Precision / Recall / F1 (non-adversary) & 0.99 / 0.98 / 0.98 \\
\midrule
\multicolumn{2}{l}{\textit{Scene-level ranking of the labeled adversary}} \\
Top-1 / Top-3 / Top-5 accuracy           & 83.0\% / 97.8\% / 99.3\% \\
\bottomrule
\end{tabular}
\end{table}

\begin{table}[H]
\centering
\caption{Reference-model comparison over 408 nuScenes scenarios. Metrics defined in Supplementary Table~\ref{si:tab:main_results}; $\bar{v}_\mathrm{col}^\mathrm{UP}$: mean collision speed (m/s) for unpreventable scenarios only.}
\label{si:tab:validity}
\renewcommand{\arraystretch}{1.25}
\footnotesize\sffamily
\setlength{\tabcolsep}{4pt}
\begin{tabular}{lccccc}
\toprule
\textbf{Reference model} & \textbf{UP\%} & \textbf{TTC$_\mathrm{min}$ $\downarrow$} & \textbf{$\bar{d}_\mathrm{min}$ $\uparrow$} & \textbf{BD$^+$\%} & \textbf{$\bar{v}_\mathrm{col}^\mathrm{UP}$ $\downarrow$} \\
\midrule
FSM            & 11.3 & 2.46 & 3.32 & 22.5 & 6.01 \\
CC-JP        & 20.3 & 0.72 & 0.83 & 56.4 & 5.06 \\
RSS       & \phantom{0}2.9 & 2.19 & 4.41 & 15.0 & 5.47 \\
\bottomrule
\end{tabular}
\end{table}

\begin{table}[H]
\centering
\caption{Full \textit{adv} trajectory kinematics. Here $j=\mathrm{d}a/\mathrm{d}t$ denotes longitudinal jerk. Subscript $\max$ denotes the per-trajectory maximum magnitude; subscript $95$ denotes the 95th percentile of the magnitude.}
\label{tab:si_kinematics}
\renewcommand{\arraystretch}{1.25}
\footnotesize\sffamily
\setlength{\tabcolsep}{3pt}
\begin{tabular}{lcccccc}
\toprule
\textbf{Method} & \multicolumn{2}{c}{\textbf{Longitudinal acceleration}} & \multicolumn{2}{c}{\textbf{Jerk}} & \multicolumn{2}{c}{\textbf{Lateral / yaw}} \\
\cmidrule(lr){2-3}\cmidrule(lr){4-5}\cmidrule(lr){6-7}
 & \textbf{$a_{\max}$} & \textbf{$a_{95}$} & \textbf{$j_{\max}$} & \textbf{$j_{95}$} & \textbf{$a_{\mathrm{lat},95}$} & \textbf{$\dot\theta_{\max}$} \\
 & (m/s$^2$) & (m/s$^2$) & (m/s$^3$) & (m/s$^3$) & (m/s$^2$) & (rad/s) \\
\midrule
\multicolumn{7}{l}{\textit{Baselines}} \\
STRIVE~\cite{rempe2022generating}  & \phantom{0}6.12 & \phantom{0}5.16 & \phantom{0}41.52 & 37.38 & \phantom{0}3.28 & 0.376 \\
SafeSim~\cite{chang2024safe}       & \phantom{0}6.61 & \phantom{0}4.68 & \phantom{0}24.90 & 16.30 & \phantom{0}1.07 & 0.317 \\
Bezier-CAT~\cite{zhang2023cat}     & 28.93 & 24.60 & \phantom{0}57.43 & 56.20 & 18.70 & 10.337 \\
\midrule
\multicolumn{7}{l}{\textit{Ours}} \\
CARS                               & \phantom{0}2.26 & \phantom{0}1.90 & \phantom{00}8.49 & \phantom{0}6.36 & \phantom{0}0.42 & 0.105 \\
CARS ($K{=}1$ \textit{adv})       & \phantom{0}3.83 & \phantom{0}3.35 & \phantom{0}18.07 & 16.14 & \phantom{0}2.13 & 0.517 \\
\midrule
\multicolumn{7}{l}{\textit{ADS-planner robustness (fixed CARS \textit{adv})}} \\
One-component diffusion planner              & \phantom{0}2.24 & \phantom{0}1.88 & \phantom{00}8.18 & \phantom{0}6.20 & \phantom{0}0.44 & 0.109 \\
CTG planner~\cite{zhong2023guided}           & \phantom{0}2.14 & \phantom{0}1.83 & \phantom{00}7.94 & \phantom{0}6.08 & \phantom{0}0.41 & 0.097 \\
\midrule
\multicolumn{7}{l}{\textit{Cross-dataset generalization}} \\
CARS on AD4CHE~\cite{zhang2023ad4che}        & \phantom{0}0.45 & \phantom{0}0.38 & \phantom{00}1.64 & \phantom{0}1.22 & \phantom{0}0.16 & 0.039 \\
CARS on RounD~\cite{krajewski2020round}      & \phantom{0}3.29 & \phantom{0}2.71 & \phantom{0}22.59 & 12.14 & \phantom{0}1.84 & 0.458 \\
\bottomrule
\end{tabular}
\end{table}

\clearpage
\section{Three Evaluation Datasets in Detail}
\label{si:datasets}

\begin{figure*}[htbp]
  \centering
  \includegraphics[width=\textwidth]{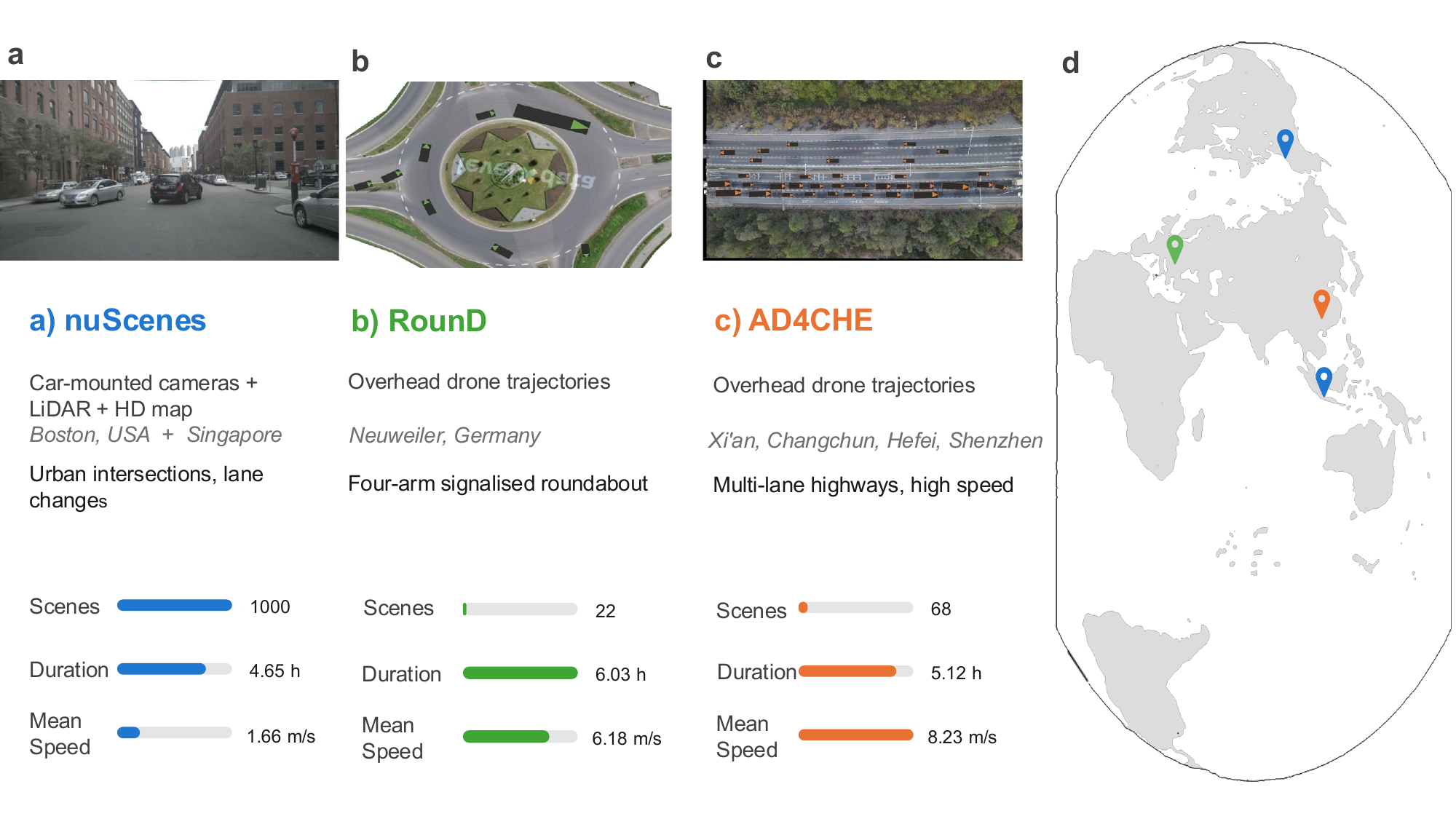}
  \caption{Three evaluation datasets spanning three continents and three road topologies.
\textbf{a}, nuScenes front-facing camera frame at a Boston (US) intersection (1{,}000 urban scenes across Boston and Singapore).
\textbf{b}, RounD overhead drone frame at the Neuweiler (Germany) roundabout (22 recordings of four-arm signalised geometry).
\textbf{c}, AD4CHE overhead drone frame on a multi-lane highway (68 recordings across four Chinese cities).
\textbf{d}, Geographic locations of the four data-collection countries; pins are colored by dataset.}
  \label{fig:si_dataset_map}
\end{figure*}

The evaluation suite spans three continents, three road topologies, and contrasting driving cultures (\Cref{fig:si_dataset_map}). nuScenes provides 1{,}000 urban scenes recorded in Boston (US) and Singapore. RounD provides 22 drone-recorded sessions of a four-arm signalised roundabout in Neuweiler, Germany. AD4CHE provides 68 drone recordings of multi-lane highway traffic across four Chinese cities. Together the three datasets stress-test CARS under signalised intersections, circular merging, and high-speed lane changing respectively.

\section{Gallery of CARS Collision Geometries on nuScenes}
\label{si:collision_gallery}

\begin{figure*}[htbp]
  \centering
  \includegraphics[width=\textwidth]{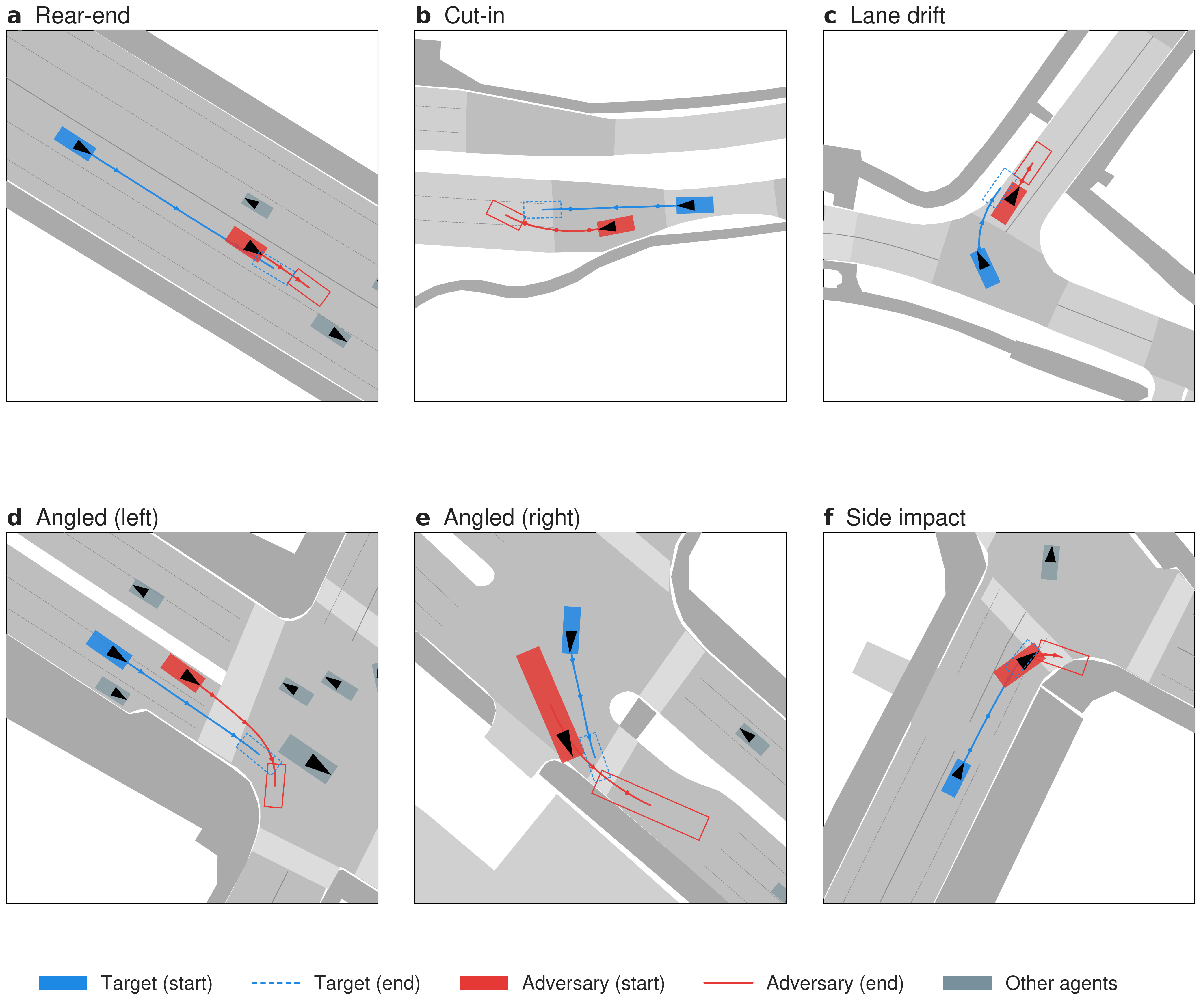}
  \caption{Gallery of collision geometries generated by CARS on nuScenes.
  \textbf{a}, Rear-end approach.
  \textbf{b}, Cut-in from ahead.
  \textbf{c}, Lane drift across the boundary.
  \textbf{d}, Angled approach from the front-left.
  \textbf{e}, Angled approach from the front-right.
  \textbf{f}, Near-perpendicular side impact.
  The \textit{tgt} is shown in \textcolor{blue}{blue} and the \textit{adv} in \textcolor{red}{red}; filled boxes mark the start position and outlined boxes the end position. Other agents are shown in gray.}
  \label{fig:si_collision_gallery}
\end{figure*}

In addition to the severity diversity quantified in main-text \Cref{tab:main_results} and the per-tier kinematic statistics in main-text Fig.~4, CARS also produces collisions across a wide variety of \emph{geometric} interaction types. \Cref{fig:si_collision_gallery} illustrates six representative cases drawn from the 408 nuScenes collision scenarios: rear-end approach, cut-in from ahead, lane drift across the boundary, two angled-approach geometries, and near-perpendicular side impact. The variety of approach geometries shows that CARS is not biased toward a single failure mode of the \textit{tgt} policy; rather, it discovers heterogeneous configurations under which the same policy fails.

\end{document}